\documentclass[lettersize,journal]{IEEEtran}
\usepackage{amsmath,amsfonts}
\usepackage{algorithmic}
\usepackage{algorithm}
\usepackage{array}
\usepackage[caption=false,font=normalsize,labelfont=sf,textfont=sf]{subfig}
\usepackage{textcomp}
\usepackage{stfloats}
\usepackage{url}
\usepackage{verbatim}
\usepackage{graphicx}
\usepackage{cite}
\usepackage{color,soul,xcolor}
\usepackage{colortbl}
\usepackage{etoolbox}
\usepackage{makecell}
\usepackage{CJKutf8}
\soulregister{\cite}7 
\soulregister{\citep}7 
\soulregister{\citet}7 
\soulregister{\ref}7 
\soulregister\eqref7
\begin{document}

\title{Air-Chamber Based Soft Six-Axis Force/Torque Sensor for Human-Robot Interaction}

\author{
{\begin{CJK}{UTF8}{bkai}
Jun Huo(霍軍), Hongge Ru(茹宏格), Bo Yang(楊博), Xingjian Chen(陳行健), Xi Li(李曦), and Jian Huang(黃劍),~\IEEEmembership{Senior Member,~IEEE}
\end{CJK}}


\thanks{J. Huo, H. Ru, B. Yang, X. Chen, X. Li and J. Huang are with the Key Laboratory of the Ministry of Education for Image Processing and Intelligent Control, School of Artificial Intelligence and Automation, Huazhong University of Science and Technology, Wuhan 430074, China.
(Corresponding author: Jian Huang, e-mail: huang\_jan@mail.hust.edu.cn).}
\thanks{Manuscript received April 19, 2021; revised August 16, 2021.}}

\markboth{IEEE TRANSACTIONS ON INSTRUMENTATION AND MEASUREMENT,~Vol.~14, No.~8, August~2021}%
{Shell \MakeLowercase{\textit{et al.}}: A Sample Article Using IEEEtran.cls for IEEE Journals}


\maketitle

\begin{abstract}
Soft multi-axis force/torque sensors provide safe and precise force interaction. 
Capturing the complete degree-of-freedom of force is imperative for accurate force measurement with six-axis force/torque sensors.
However, cross-axis coupling can lead to calibration issues and decreased accuracy. 
In this instance, developing a soft and accurate six-axis sensor is a challenging task. 
In this paper, a soft air-chamber type six-axis force/torque sensor with 16-channel barometers is introduced, which housed in hyper-elastic air chambers made of silicone rubber. 
Additionally, an effective decoupling method is proposed, based on a rigid-soft hierarchical structure, which reduces the six-axis decoupling problem to two three-axis decoupling problems. Finite element model simulation and experiments demonstrate the compatibility of the proposed approach with reality. 
The prototype's sensing performance is quantitatively measured in terms of static load response, dynamic load response and dynamic response characteristic.
It possesses a measuring range of 50 N force and 1 Nm torque, and the average deviation, repeatability, non-linearity and hysteresis are 4.9$\%$, 2.7$\%$, 5.8$\%$ and 6.7$\%$, respectively.
The results indicate that the prototype exhibits satisfactory sensing performance while maintaining its softness due to the presence of soft air chambers.
\end{abstract}

\begin{IEEEkeywords}
soft sensor, six-axis force/torque sensor, air-chamber type sensor, silicone rubber.
\end{IEEEkeywords}

\section{Introduction}
\IEEEPARstart{I}{n} 
the field of human-robot force interaction, ensuring safety is of utmost importance. 
Rigid sensors can create an opposing instant reaction force on the human body, which is not favorable. 
Therefore, incorporating softness into the physical design is an effective strategy to ensure safe interaction \cite{SciroThuruthel}. 
Soft force sensors are crucial in connecting the robot and its surroundings, detecting interaction forces, and reducing the impact of shocks \cite{ref18}. 
Consequently, soft sensors are indispensable for developing reliable and robust control systems for human-robot interaction. 
Additionally, they are preferred in human-robot interaction to promote a more friendly and positive interaction \cite{RALmis}.

In order to accurately measure the state of interaction, sensors must possess both a soft structure and the ability to measure multi-axial interaction force, including its direction and magnitude. 
Six-axis force/torque sensors (abbreviated as 6-F/T sensors) can measure the full degree-of-freedom of a force or torque applied to a point in space. 
These sensors have proven to be valuable in various robotics research areas, including dexterous manipulation using robot hands \cite{KimTRO,BoTMRB}, tactile sensing \cite{refDw}, surgical robotics \cite{refNoh}, humanoid robots \cite{OzawaTRO} and bio-mechanical applications \cite{Yan,refLong}. 
However, current research primarily focuses on the rigid six-axis sensor, leaving a significant gap for the development of soft six-axis sensors that would be applicable to human-robot interaction.

Soft materials exhibit a non-linear property that diminishes measuring accuracy.
So it is challenging to design and fabricate a 6-F/T sensor that excels in soft, accuracy and wide measurement range. 
Many attempts have been investigated to develop soft, compatible, and multi-dimensional force sensors with traditional sensing technologies, such as resistance strain gauges \cite{Maita,ref41}, piezoelectric \cite{ref42,ref43,ref45}, capacitive \cite{DHyuk,ref12,ref27}, photoelectric  \cite{ref1}, and soft magnetic technologies \cite{ref48,refXie}. 
These types of sensors are usually limited by hysteresis, drift over time, short lifetime, slow response, and multiple complex manufacturing methods \cite{TAWK}.
Air-chamber type sensors based on air-pressure sensors are widely used in soft multi-axial force measurement due to their softness, adjustable measuring range, high accuracy, and customizable size \cite{Choi}.
The air-chamber type 6-F/T sensor based on vulcanizing silicone rubber is an great application for balancing softness and accuracy.
For instance, Kong et al. \cite{ref37} developed an insole force sensing pad that used silicone tubes and pressure sensors to measure ground reaction force (GRF). 
Jacobs et al. and Rossi et al. \cite{ref38,ref39} introduced elastic materials with pressure sensors and optical sensors to measure GRF, respectively. 
Choi et al. \cite{Choi} proposed a soft three-axis force sensor using radially symmetric pneumatic chambers.
Owing to the complexity of design and manufacturing 6-F/T sensor, the above study only 
measured GRF, losing the applied torque information.
Gong et al. \cite{ref46} developed a pneumatic tactile sensor applied to a robotic fingertip.
Yang et al. \cite{Yang} fabricated a silicone pneumatic soft sensor in a soft robotic gripper based on conventional molding techniques, which can measure 1.5 N contact force.
Vogt et al. \cite{Vogt} developed a soft multi-axial force sensor using elastic materials and liquid metals to measure 5 N normal force and 1 N shear force. 
These soft, accurate sensors are limited by measurement range.
Although methods for measuring a larger range of forces with soft sensors were developed \cite{ref50,ref51}, they encountered difficulties to ensure the linearity in the sensor measurements.
One solution to optimize measurement range and accuracy, as well as improve sensor linearity, is to increase the number of channels.
However, this also results in a larger sensor size and a more complex manufacturing process, which conflicts with the desired small and lightweight design and makes it difficult to arrange air chambers.
To address these issues, a rigid-soft hierarchical multi-material structure is proposed as a cost-effective solution that can provide both softness and high accuracy performance simultaneously.

Coupling between axes makes calibrating the 6-F/T sensor complicated and inaccurate \cite{Akbari}. 
Because of hysteresis and time varying \cite{rhgMMT}, coupling interference will be even more intense in soft materials.
Hence, how to obtain a high-precision decoupling method presents a challenge. 
Several prior work has been done on decoupling multi-dimensional force/torque. 
For example, Kim et al. \cite{ref40} developed an optical sensor-based multi-dimensional force sensing unit.
A linear transformation was established to figure out the relationship between the photointerrupter signals and 3-axial force.
Zhao et al. \cite{ZhaoY} introduced a rigid-flexible hybrid parallel three-dimensional force sensor that used a static mathematical model based on the vector method to obtain multi-axis force. 
Wang et al. \cite{ref16} proposed a decoupling method based on a designed Wheatstone bridge circuit, the fingerprint-patterned micro-fluidic channels and the top oval-shaped protrusion. 
Previous major decoupling approaches are based on empirical linear hypotheses, 
while potential rules is not clear, resulting in the uncertain structure of the empirical model.
A large amount of calibration data is needed to fit the complex structure of the model.
In this case, a linearized theoretical model and a simple conversion law between sensor measurements and the six-axis force component should be constructed \cite{ref22}.  
To address this issue, this paper proposes a model-based theoretical decoupling method. 
The linear relationship between sensor unit deformation and loading force/torque is found by the sensing mechanism model of a single sensor unit.
The sensing model-based decoupling strategy has fewer calibration parameters and a clear physical meaning.

In this paper, a soft air-chamber type 6-F/T sensor is proposed for human-robot interaction application. The prototype consists of a hard shell layer and a soft layer, each with 8 air cavities, for a total of 16 channels.
The volume change of a single air chamber is analyzed when subjected to uniaxial force loading. 
Furthermore, the corresponding pressure change of the air chamber is obtained.
Since each axis exhibits a linear characteristic, a decoupling method is proposed based on the rigid and flexible layered structure to determine the conversion law between pressure and loading force/moment. 
The contributions of the paper are summarized as follows.
\begin{enumerate}
\item{A novel soft six-axis force/torque sensor based on the air chamber is designed and manufactured. The proposed sensor have 16 channels, distributed in the soft upper layer and the rigid lower layer.
This configuration extends the measurement range through increasing  channels while preserving accuracy, making it suitable in human-contact scenarios.
}
\item{A six-dimensional force/torque decoupling method is proposed based on the principle analysis of the air-chamber type sensor and the rigid-soft hierarchical structure. 
To the best of our knowledge, in accordance with different forms of stress, the implementation of a layered structure, comprising of rigid and soft materials, is firstly applied to the six-dimensional force/torque decoupling technique.
Two sets of tri-axial force/torque are divided by the rigid-soft hierarchical shell, reducing the six-axis decoupling problem to two three-axis decoupling problems.
The number of calibration parameters is reduced from 96 to 6, which simplifies the coupling complexity.}
\item{The validity of the prototype is confirmed by both finite element modeling (FEM) simulations and experiments.
The prototype offers the benefit of being soft while performing at the same level as other sensors.}
\end{enumerate}

The remainder of this paper is organized as follows. The design, fabrication of the sensor unit  are introduced in section II. The principle analysis of the six-axis air-chamber type sensor as well as finite element model analysis are introduced in section III. Pilot study is presented in section IV. Discussions and conclusion are in section V and section VI separately.

\section{Design and Fabrication of the Sensor Unit}
In this section, the proposed six-axis force/torque sensor based on air pressure sensing is introduced.
\subsection{Mechanical Design and Fabrication of the Sensor Unit}
The proposed sensor is designed for sensing six-axis force or torque in human-robot interaction applications, specifically as a wearable sensor. 
It necessitates a high level of softness to prevent potential damage during interactions while concurrently maintaining a large measurement range to accommodate abrupt shocks.
Silicone rubber is considered the most suitable material for achieving both softness and accurate measurement. 
Vulcanizing silicone rubber (Elastosil M4601, Wacker Chemie AG, Germany) has been widely used in the production of soft robots due to its high elongation and good tear strength \cite{refAlici,refHuangX}.

\begin{figure}[!t]
\centering
\includegraphics[width=3.2in]{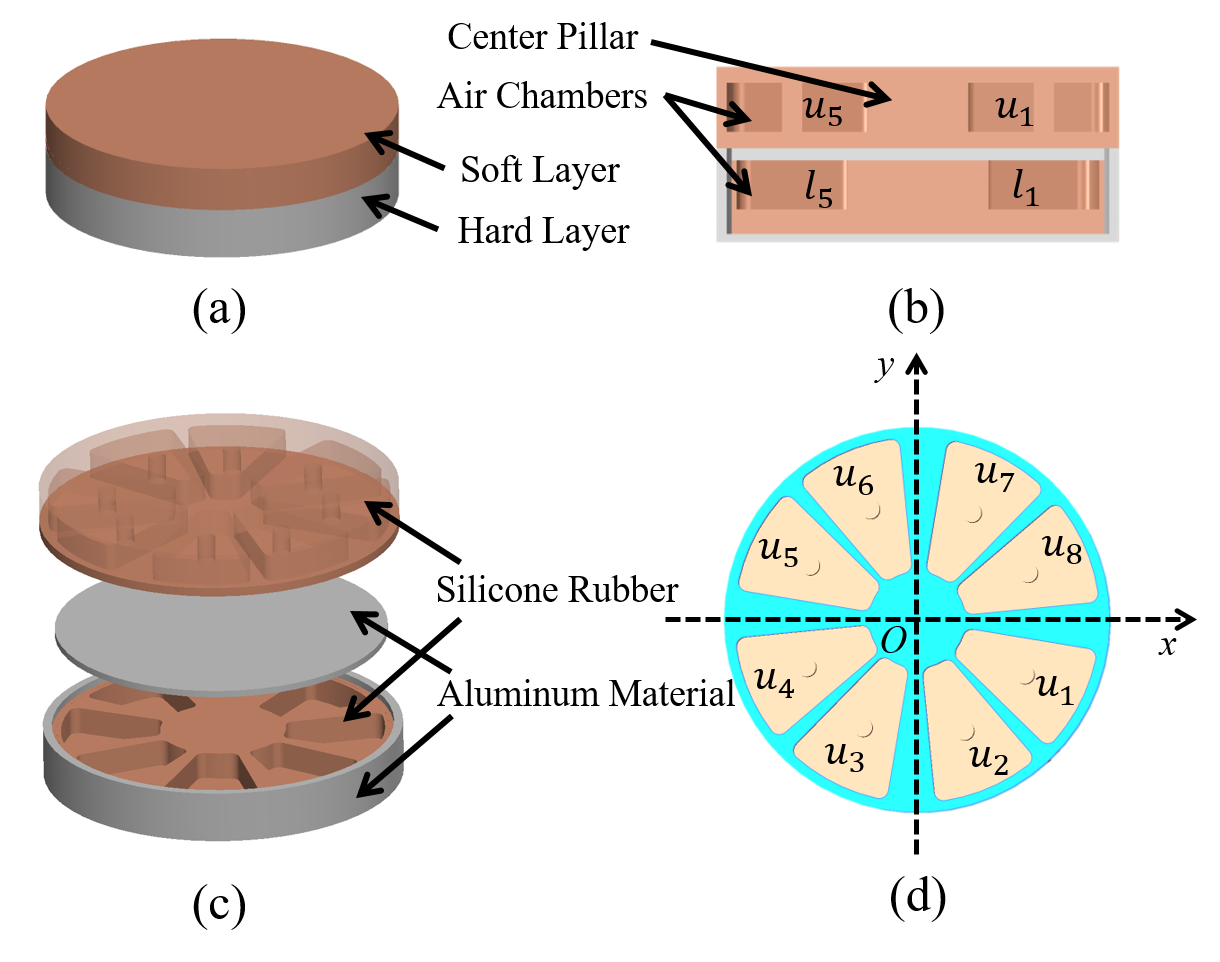}
\caption{The overview of the proposed sensor. (a) The proposed prototype model. (b) The front section view of the sensor model. (c) The explosion view of the sensor model. (d) The top section view of the upper layer sensor model. The numbers refer to the air chambers, $u$ donates the upper layer, $l$ donates the lower layer.}
\label{sensorview}
\end{figure}

Fig.~\ref{sensorview} provides an overview of the proposed sensor, comprising two layers, each featuring eight air chambers equipped with absolute barometers.
The upper layer is made entirely of silicone rubber, while the lower layer is reinforced with a rigid aluminum alloy (AA) material, to increase rigidity.
The incorporation of rigid materials can achieve not only the segregation of force directly but also enhance the precision of measurement. 
The eight air chambers in the upper layer are distributed to sense z-axis torque in pairs, while the eight air chambers in the lower layer are evenly distributed.

\begin{figure*}[!t]
\centering
\includegraphics[width=1.0\textwidth]{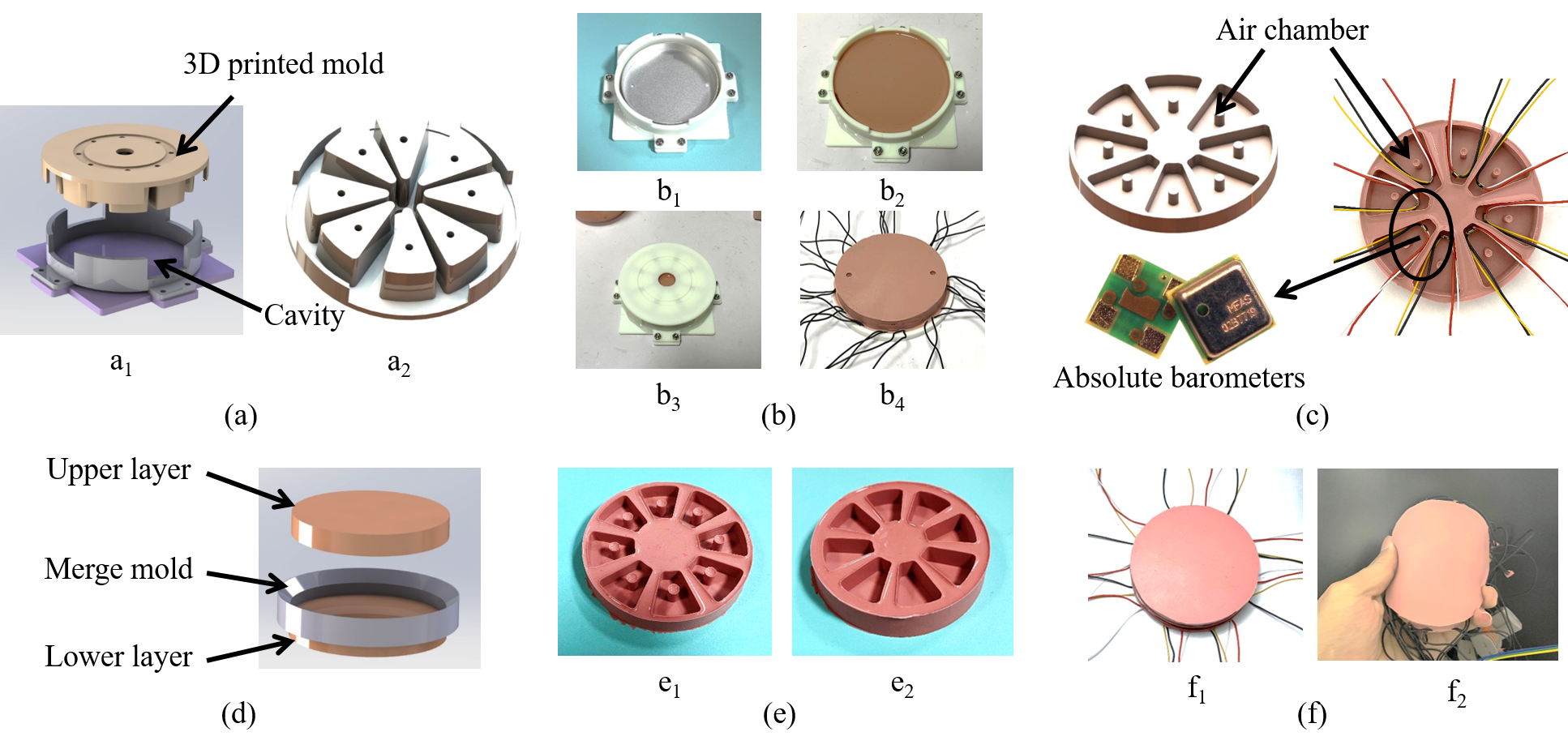}
\caption{Fabrication process of the proposed six-axis sensor. (a) Step 1, fabricate the upper soft layer. $\rm a_1$, 3D printed mold, pour the silicone rubber into the cavity of the molding. $\rm a_2$, the enlarged cavity mold.
(b) Step 2, fabricate the lower hard layer. $\rm b_1$, hard layer mold. $\rm b_2$, pour liquid silicone rubber. $\rm b_3$, cover the lid. $\rm b_4$, embedded the barometers and seal the bottom.
(c) The internal image with absolute barometers.   
(d) Step 3, use mold to merge the upper layer and the lower layer.
(e) The internal images without barometers. $\rm e_1$, upper soft layer. $\rm e_2$, lower hard layer.
(f) The prototype ($\rm f_1$) and the bendable characteristics display of the soft upper layer ($\rm f_2$).}
\label{fabrication}
\end{figure*}

The fabrication process of the proposed sensor is outlined in Fig.~\ref{fabrication}, comprising three key steps. 
In the first step, the base of the two air cavities is formed using silicone rubber. The upper soft cavity is entirely made of silicone rubber, as depicted in Fig.~\ref{fabrication}(a). Conversely, the lower layer is manufactured by pouring silicone rubber into a fixed rigid AA shell, as Fig.~\ref{fabrication}(b).
The second stage involves inserting the barometer into the air chamber and affixing it securely with adhesive, as illustrated in Fig.~\ref{fabrication}(c).
Following this, the cavity's base is encapsulated using the mold, and the two parts are merged to finalize the fabrication process, as depicted in Fig.~\ref{fabrication}(d).
Fig.~\ref{fabrication}(e)(f) is the prototype and shows its soft characteristics.

The proposed sensor's size is determined by the air chamber, with a designed diameter of 80 mm and a thickness of 20 mm. The overall diameter can be reduced by adjusting the outer wall thickness and air chamber size since the air chamber's size is tunable. 
The measurement range is determined based on the proportion of the chamber to solid silicone, with a wider range achievable through a larger proportion of solid silicone. While, sensitivity and measuring range are contradictory, with sensitivity decreasing as the measuring range expands.

\subsection{Communication Setup}
Each air chamber is equipped with an individual barometer, as shown in the Fig.~\ref{fabrication}(c). 
An absolute barometer (MS5637-02BA03, TE Connectivity Measurement Specialties) is used to measure the air pressure of the air chamber. This barometer operates within the range of 10 to 2000 mbar over a temperature span of -40 to 85 $^\circ$C, offering a resolution of 0.039 mbar.
The barometer features an analog-to-digital converter and an IIC ($\rm I^2C$) communication protocol, allowing direct measurement of the air pressure without any electrical peripherals. To communicate with the computer, we use a Micro Control Unit (Arduino Mega 2560). The sample rate of the prototype is set to 1024 $\rm Hz$.

\section{Decoupling Methodology of the Six-Axis Air-Chamber Type Sensor}
\subsection{Measurement Method}
\begin{figure}[!t]
\centering
\includegraphics[width=3.2in]{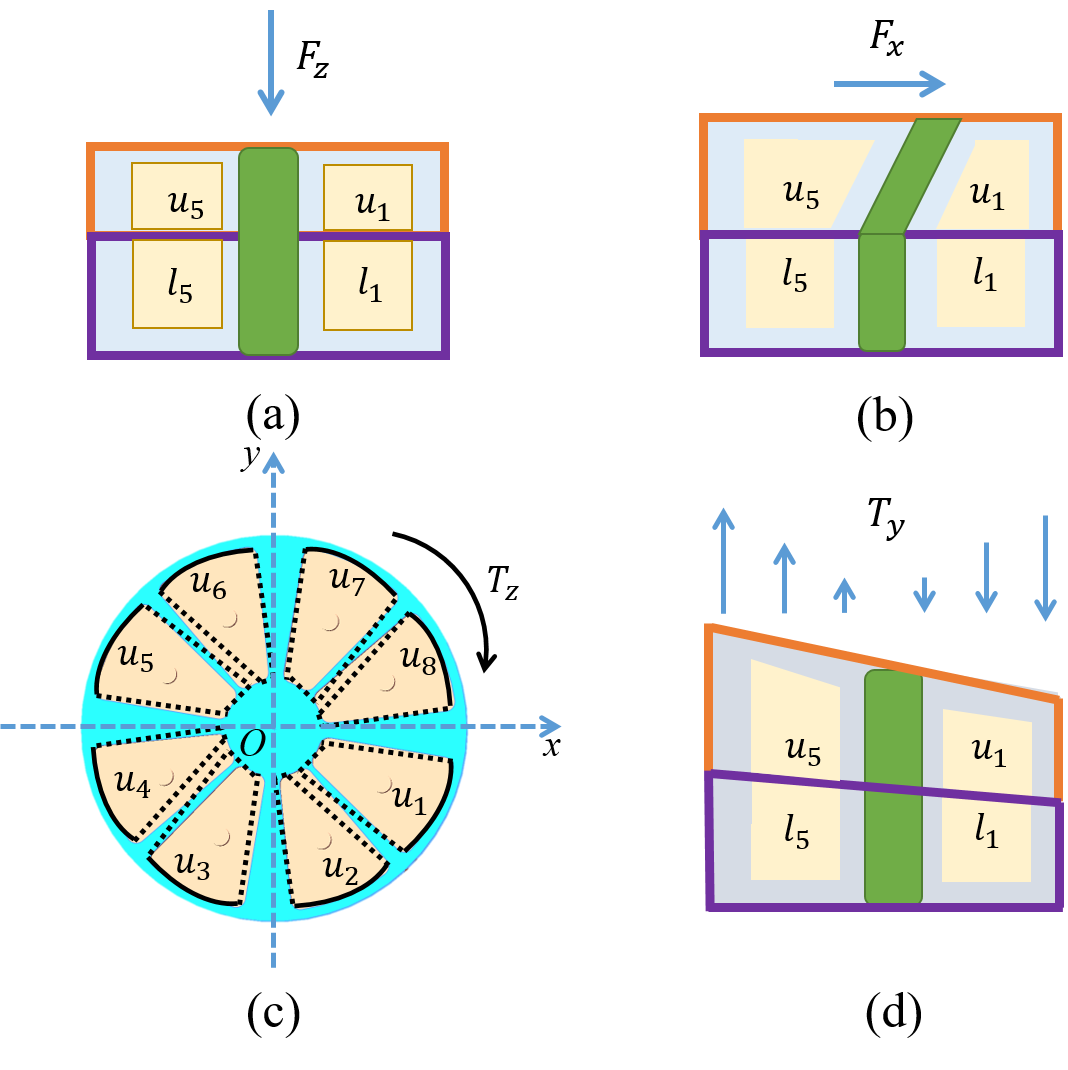}
\caption{Cross-sectional area deformation analysis of pneumatic chambers. (a) Case I, applied single axis force $F_z$, the upper layer and the lower layer are all compressed. (b) Case II, applied single axis force $F_x/F_y$, the upper layer slips along with the direction of the force, the lower layer makes no change. (c) Case III, applied single axis torque $T_z$, the upper layer rotates along with z-axis (the dotted line shows the deformation), the lower layer makes no change. (d) Case IV, applied single axis torque $T_x/T_y$, one side is compressed and the other side is stretched.}
\label{fig_FT}
\end{figure}

The volumetric changes in compression differ between the upper and lower layers.
Fig.~\ref{fig_FT} illustrates the deformation analysis of pneumatic chambers' cross-sectional areas based on our sensor configuration. 
We utilize both upper and lower layers to detect distinct Force/Torque dimensions. Specifically, the lower layer responds exclusively to pressure $F_z$ and torque $T_x$/$T_y$, whereas the upper layer is highly sensitive to all dimensions. 
Notably, the cross-sectional shape is circular, without loss of generality, we analyze the effects of $F_x$ and $T_y$. 
Consequently, the lower cavity structure is employed to gauge normal stress $F_z$, torque $T_x$ and $T_y$, whereas the upper layer senses all six-axis force/torque components, allowing for the separation of shear stress $F_x$/$F_y$ and torque $T_z$,
which is,
\begin{align}
\label{FT1}
\mathbf p_l &= \mathbf f_l(F_z, T_x, T_y),\\
\begin{split}
\mathbf p_u &= \mathbf f_{u1}(F_x, F_y, T_z)
 + \mathbf f_{u2}(F_z, T_x, T_y)\\
& = \mathbf f_u(F_x, F_y, F_z, T_x, T_y, T_z),
\end{split}
\end{align}
where, $F_x, F_y$ and $F_z$ are the applied force on x-axis, y-axis and z-axis respectively, $T_x, T_y$ and $T_z$ are the applied torque on x-axis, y-axis and z-axis, respectively.
$\mathbf p_l=[p_{l1}, \,p_{l2}, \,p_{l3}, \,p_{l4}, \,p_{l5}, \,p_{l6}, \,p_{l7}, \,p_{l8}]^T$, $\mathbf p_u=[p_{u1}, \,p_{u2}, \,p_{u3}, \,p_{u4}, \,p_{u5}, \,p_{u6}, \,p_{u7}, \,p_{u8}]^T$, are the air pressure vector of the lower layer air chambers and the upper layer air chambers respectively.
$\mathbf f_l, \mathbf f_u$ are the lower layer function relationship and the upper layer function relationship between applied force/torque and the air pressure vector. 

\begin{figure}[!t]
\centering
\includegraphics[width=3.5in]{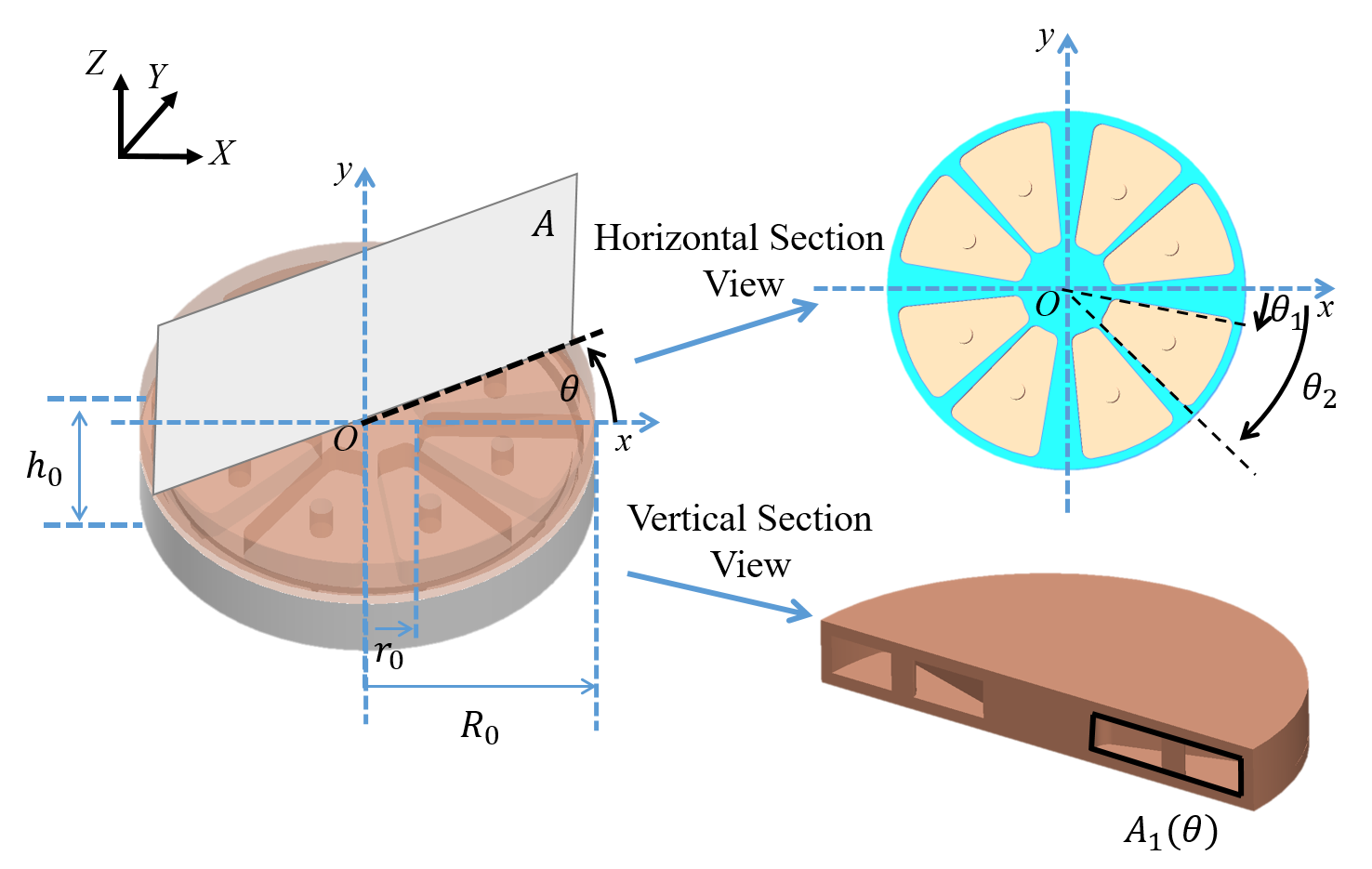}
\caption{Cross-sectional area of the air chamber model analysis defined by $A_1(\theta)$.
A cylindrical column at the center of each chamber is used to support the cavity and prevent excessive collapse.}
\label{fig_cham}
\end{figure}

\subsection{Chamber Deformation Analysis}
In this section, we analyze the overall volume change of the chambers. The cross-section area of the chamber can be assumed as a rectangle, as shown in Fig.~\ref{fig_cham}.
Then the volume of single chamber can be regarded as an integral of multiple rectangles, which is calculated as follows,
\begin{equation}
\label{Ch1}
V=\int^{\theta_2}_{\theta_1} A_1(\theta)rd\theta,
\end{equation}
where, $r=\frac{1}{2}(R_0+r_0)$. $R_0, r_0$ are the radius of the prototype sensor and the center pillar, respectively.
$\theta_1, \theta_2$ are the position of the air chambers, as shown in Fig.~\ref{fig_cham}.
Then we consider the volume change of 8 chambers in a single layer under only one interaction on the basis of Eq.~\ref{Ch1}. 
\begin{figure}[!t]
\centering
\includegraphics[width=3.2in]{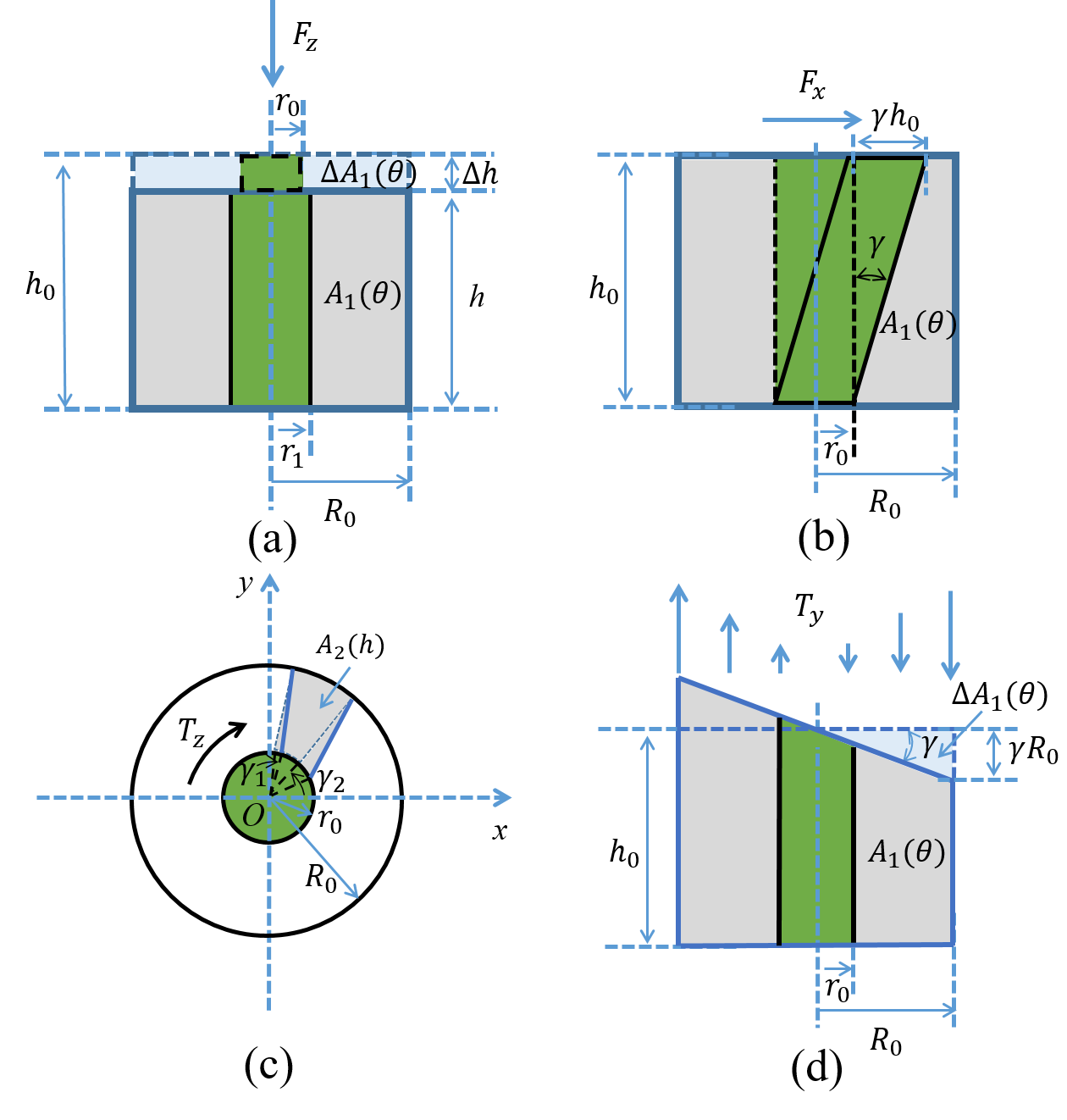}
\caption{The single air chamber deformation analysis. Green area represents the central pillar and gray area refers to the air chamber.
(a) Case I, $F_z$ applied. (b) Case II, $F_x/F_y$ applied. (c) Case III, $T_z$ applied. (d) Case IV, $T_x/T_y$ applied.}
\label{fig_FT2}
\end{figure}

\subsubsection{Volume change under a normal force Fz}
The section view of the chamber under the applied $F_z$ is shown as Fig.~\ref{fig_FT2}(a). The area of $A_1(\theta)$ is,
\begin{equation}
\label{Fz1}
A_1(\theta)=h(R_0-r_1),
\end{equation}
where, $r_1$ is the new radius of the center pillow after deformation. According to mechanics of materials \cite{refDowling}, the deformation relations after loading are as follows,
\begin{align}
\label{Fz2}
h&=h_0(1-\epsilon_{zz}),\\
r_1&=r_0(1-\epsilon_{zx}),
\end{align}
where, $\epsilon_{zz}$ and $\epsilon_{zx}$ refer to the normal strain and the shear strain, which can be obtained as below,
\begin{align}
\label{Fz3}
\epsilon_{zz}&=\frac{1}{{\rm E}S}F_z,\\
\epsilon_{zx}&=-\frac{\nu}{{\rm E} S}F_z,
\end{align}
where, $\rm E$ is Young's modulus of the rubber material, $S$ is the cross-section area perpendicular to normal stress of the centre pillar. $\nu$ is the Poisson ratio of the material.

Substitute Eq.~\ref{Fz2} to Eq.~8 into Eq.~\ref{Fz1}, we can obtain
\begin{equation}
\begin{split}
\label{Fz4}
A_1(\theta)&=h_0(1-\epsilon_{zz})(R_0-r_0(1-\epsilon_{zx}))\\
&=h_0(R_0-r_0)-\epsilon_{zz}R_0h_0+\epsilon_{zx}r_0h_0-\epsilon_{zz}\epsilon_{zx}r_0h_0.
\end{split}
\end{equation}

Ignore higher-order terms, which assumes $\epsilon_{zz}\epsilon_{zx}=0$, so the change of $A_1(\theta)$, $\Delta A_1(\theta)$ could be simplified as Eq.~\ref{Fz5},
\begin{equation}
\begin{split}
\label{Fz5}
\Delta A_1(\theta)&=-\frac{h_0R_0}{{\rm E}S}F_z+\frac{-\nu r_0h_0}{{\rm E}S}F_z\\
&=-\frac{h_0}{{\rm E}S}(R_0+\nu r_0)F_z.
\end{split}
\end{equation}

According to Eq.~\ref{Ch1}, because the eight air chambers in the lower layer are evenly distributed, see Fig.~\ref{fig_cham}, take $\theta_2-\theta_1=\frac{\pi}{8}$, we can obtain the volume change $\Delta V_k$, which is
\begin{equation}
\label{Fz_n1}
\Delta V_k=-\frac{\pi rh_0}{8{\rm E}S}(R_0+\nu r_0) F_z,
\end{equation}
$k$ is the label of the air chamber, then, the volume change of eight lower air chambers can be written as,
\begin{align}
\label{Fz_n}
\mathbf{\Delta V} &= -\alpha 
\mathbf T_{Fz}
F_z,\\
\alpha&=\frac{\pi rh_0}{8{\rm E}S}(R_0+\nu r_0),
\end{align}
where, $F_z$ is a scalar quantity, $\alpha$ is a constant, $\mathbf T_{Fz}=[1\;1\; 1\; 1\; 1\; 1\; 1\; 1]^T$, $\mathbf{\Delta V} = [\Delta V_1,\Delta V_2,\Delta V_3,\Delta V_4,\Delta V_5,\Delta V_6,\Delta V_7,\Delta V_8]^T$. 
The relationship formula between volume change and applied force demonstrates linearity.
\subsubsection{Volume change under a shear force $F_x/F_y$}
\begin{figure}[!t]
\centering
\includegraphics[width=3.4in]{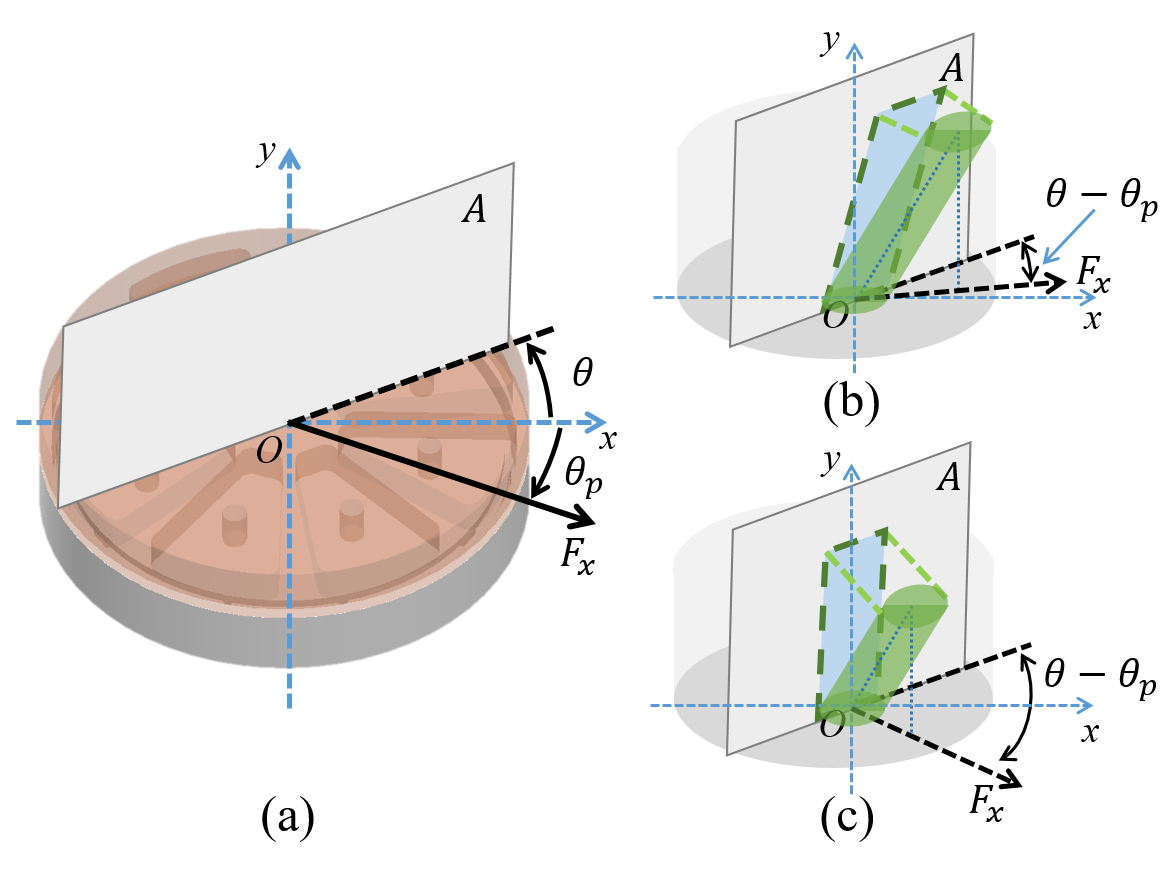}
\caption{The sensor deformation analysis under the shear force $F_x$. (a) The definition of shear force angle $\theta_p$ and cross-section angle $\theta$. (b) In the case of $\theta-\theta_p=0$, the center pillar is parallel to the plane $A$, and the projection of the center pillar has a maximum deformation. (c) In the case of $\theta-\theta_p=\frac{\pi}{2}$, the center pillar is vertical to the plane $A$, and the projection of the center pillar has a minimum deformation.}
\label{cham_Fx}
\end{figure}
In this case, assume that the applied force is concentrated at the centre, so the edge deformation is ignored.
According to Fig.~\ref{cham_Fx}, 
the blue part of the cross section $A$ represents the projection of the central pillar onto plane $A$, for example, when $\theta-\theta_p=0$, as shown in Fig.~\ref{cham_Fx}(b), the stretching direction of the central pillar is parallel to plane $A$, resulting in maximum deformation in the projection of the central pillar onto plane $A$.
When $\theta-\theta_p=\frac{\pi}{2}$, illustrated in Fig.~\ref{cham_Fx}(c), the stretching direction of the central pillar is perpendicular to plane $A$, and the projection of the central pillar on the plane $A$ does not change.
Then the cross section $A_1(\theta)$ is calculated as Eq.~\ref{Fx2}, according to Fig.~\ref{fig_FT2}(b),
\begin{equation}
\label{Fx2}
A_1(\theta)=(R_0-r_0)h_0-\frac{1}{2}\gamma h_0^2{\rm cos} (\theta-\theta_p),
\end{equation}
where, $\theta_p$ is the reference direction of the force $F_x$, $\gamma$ is the shear strain resulted by the shear stress, see Fig.~\ref{fig_cham}. Then the area change can be written as
\begin{equation}
\label{Fx3}
\Delta A_1(\theta)=\frac{1}{2}\gamma h_0^2{\rm cos} (\theta-\theta_p) \\
=\frac{1}{2}\gamma h_0^2({\rm cos} \theta {\rm cos} \theta_p + {\rm sin} \theta {\rm sin} \theta_p),
\end{equation}
according to reference \cite{refDowling}, the shear strain $\gamma$ can be calculated as
\begin{equation}
\label{Fx4}
\gamma=\frac{2(1+\nu)}{\rm E}\sigma_{p},
\end{equation}
substitute Eq.~\ref{Fx4} into Eq.~\ref{Fx3}, we can obtain
\begin{equation}
\begin{split}
\label{Fx5}
\Delta A_1(\theta)&= \frac{(1+\nu)h_0^2}{{\rm E}S_p}({\rm cos} \theta {\rm cos} \theta_p + {\rm sin} \theta {\rm sin} \theta_p)\mathbf F \\
&=\frac{(1+\nu)h_0^2}{{\rm E}S_p}
\begin{bmatrix} 
{\rm cos} \theta & {\rm sin} \theta
\end{bmatrix}
\begin{bmatrix} 
F_x\\F_y
\end{bmatrix},
\end{split}
\end{equation}
where, $S_p$ refers to the cross-section area perpendicular to shear stress of the centre pillar.
Then, we can obtain the volume change of the eight pneumatic chambers according to Eq.~\ref{Ch1}, which is
\begin{align}
\label{Fx_1}
\mathbf{\Delta V}&=\beta 
\mathbf T_{xy}
\begin{bmatrix} 
F_x \\ F_y
\end{bmatrix},\\
\beta &= \frac{(1+\nu)rh_0^2}{{\rm E}S_p},
\end{align}
where, $\beta=$ is a constant. The matrix $\mathbf T_{xy}$ is shown in the supplementary material.
Therefore, the relationship between volume change and applied force $F_x/F_y$ is linear. 

\subsubsection{Volume change under a torque $T_z$}
In this condition, compression affects only the upper layer chambers.
Unlike other situations, volume change is to be considered in the vertical direction. 
The shear strain model is as shown in Fig.~\ref{fig_FT2}(c). Assume that the volume of the air chamber varies uniformly.
$A_2(h)$ represents cross-section area, which is
\begin{equation}
\label{Mz1}
A_2(h)=\frac{1}{8}\pi(R_0^2-r_0^2)+\frac{1}{2}(R_0-r_0)r_0(\gamma_2-\gamma_1),
\end{equation}
where the strain $\gamma$ can be expressed as
\begin{align}
\label{Mz2}
\gamma_1 &=\frac{1}{{\rm G_1}S_1}T_z,\\
\gamma_2 &=\frac{1}{{\rm G_2}S_2}T_z,
\end{align}
where, $\rm G_1, G_2$ are the shear modulus, $S_1, S_2$ refer to the shear stress surface.
since the volume is integrated vertically, substitute Eq.~\ref{Mz2} and Eq.~22 into Eq.~\ref{Mz1}, we can obtain
\begin{align}
\label{Mz3}
\begin{split}
\Delta V_k &= \int^{h_0}_0 \frac{h}{h_0}\Delta A_2(h)dh\\
&=\frac{1}{4}(R_0-r_0)r_0h_0(\gamma_2-\gamma_1),\\
\end{split}
\\ &\mathbf{\Delta V}= \xi \mathbf T_{Tz} T_z,
\\ &\xi=\frac{1}{4}(R_0-r_0)r_0h_0(\frac{1}{{\rm G_2}S_2}-\frac{1}{{\rm G_1}S_1}),
\end{align}
where, $\mathbf T_{Tz}=[1,\; -1,\; 1,\; -1,\; 1,\; -1,\; 1,\; -1]^T$, $\xi$ is a constant. Hence, the relationship between volume change and the applied torque 
$T_z$ exhibits linearity.

\subsubsection{Volume change under a torque $T_x/T_y$}
In this case, assume that the lower layer deforms into an inclined plane, 
according to Fig.~\ref{fig_FT2}(d), similar to case $F_x/F_y$, the cross section $A_1(\theta)$ can be expressed as 
\begin{equation}
\label{Tx1}
A_1(\theta)=(R_0-r_0)h_0-\frac{1}{2}\gamma (R_0^2-r_0^2){\rm cos} (\theta-\theta_p),
\end{equation}
where, $\theta_p$ is the reference direction of the torque $T_x/T_y$, which is defined as perpendicular to the rotating axis. 
Then the area change can be written as
\begin{equation}
\label{Tx2}
\begin{split}
\Delta A_1(\theta)&=\frac{1}{2}\gamma (R_0^2-r_0^2){\rm cos} (\theta-\theta_p)\\
&=\frac{1}{2}\gamma (R_0^2-r_0^2)({\rm cos} \theta {\rm cos} \theta_p+{\rm sin} \theta {\rm sin} \theta_p).
\end{split}
\end{equation}

The parameter $\gamma$ can be expressed as
\begin{equation}
\label{Tx3}
\gamma=\frac{h_0\epsilon_p}{R_0}=\frac{h_0}{{\rm E} S_pR_0^2}\mathbf{T}.
\end{equation}

Substitute Eq.~\ref{Tx3} into Eq.~\ref{Tx2}, we can obtain that
\begin{equation}
\label{Tx4}
\Delta A_1(\theta)=\frac{h_0}{2{\rm E}S_pR_0^2}(R_0^2-r_0^2)
\begin{bmatrix} 
{\rm cos} \theta & {\rm sin} \theta
\end{bmatrix}
\begin{bmatrix} 
T_x\\T_y
\end{bmatrix}.
\end{equation}

Then, we can obtain the volume change of the air chambers according to Eq.~\ref{Ch1}, which is
\begin{align}
\label{Tx6}
\mathbf{\Delta V}&=\lambda \mathbf T_{xy}
\begin{bmatrix} 
T_x\\T_y
\end{bmatrix},\\
\lambda&=\frac{rh_0}{2{\rm E}S_pR_0^2}(R_0^2-r_0^2),
\end{align}
where, $\lambda$ is a constant, so the relationship between the volume change and applied torque $T_x/T_y$ can be considered as linear characteristic.

\subsubsection{Volume change under multi-dimensions force/torque}
The above four cases have proved that the volume change and pressure is linear characteristic. One of the characteristics of linearity is additive.
The transition law between the applied force/torque and the volume change is as follows, 
\begin{align}
\label{MT1}
\mathbf{\Delta V}_l&=\mathbf T_l
\begin{bmatrix} 
F_z\\T_x\\T_y
\end{bmatrix},\\
\mathbf{\Delta V}_u &=
\mathbf T_{u1}
\begin{bmatrix} 
F_x\\F_y\\T_z
\end{bmatrix}
+\mathbf T_{u2}
\begin{bmatrix} 
F_z\\T_x\\T_y
\end{bmatrix},
\end{align}
where, $\mathbf{\Delta V}_l$ is volume change of the lower layer chambers, $\mathbf{\Delta V}_u$ is the volume change of the upper layer chambers. 
The coefficient matrix are as follows, 
where $\alpha, \beta, \xi, \lambda$ are defined as Eq.~13, Eq.~19, Eq.~25 and Eq.~31,
the subscript $(\cdot)_u, (\cdot)_l$ refer to the parameters mentioned above of the upper layer and lower layer,
\begin{align}
\label{MT3}
\mathbf{T_{l}} &=
\begin{bmatrix} 
-\alpha_l \mathbf T_{Fz} &\lambda_l \mathbf T_{xy}
\end{bmatrix},\\
\mathbf T_{u1} &=
\begin{bmatrix}
    \beta_u \mathbf T_{xy} & \xi_u \mathbf T_{Tz}
\end{bmatrix},\\
\mathbf T_{u2} &=
\begin{bmatrix}
    -\alpha_u \mathbf T_{xy} & \lambda_u \mathbf T_{xy}
\end{bmatrix}.
\end{align}

\begin{figure}[!t]
\centering
\subfloat[]{\includegraphics[width=2.2in]{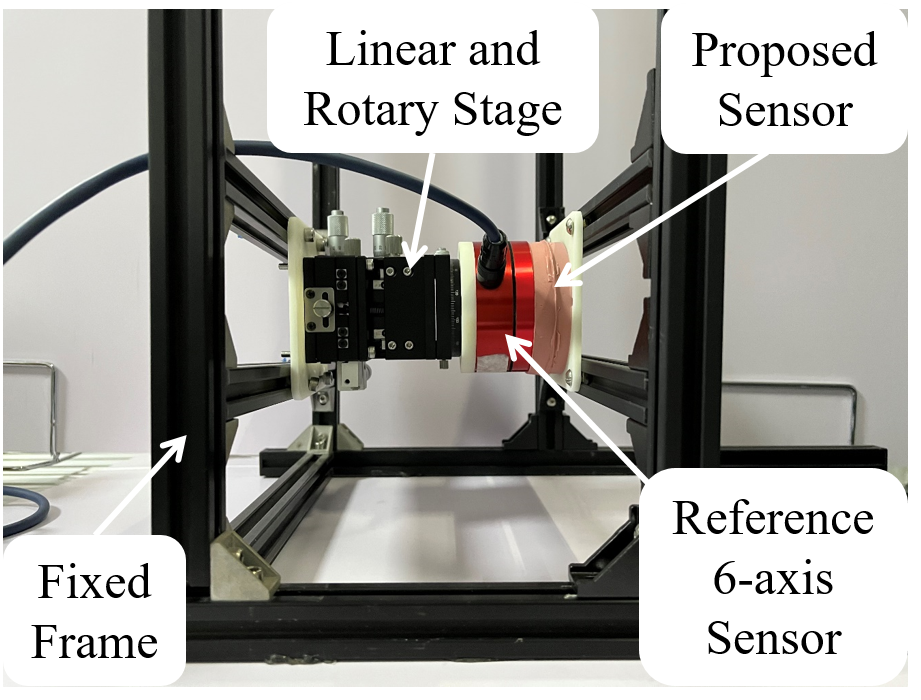}
\label{a}}
\hfil
\subfloat[]{\includegraphics[width=2.2in]{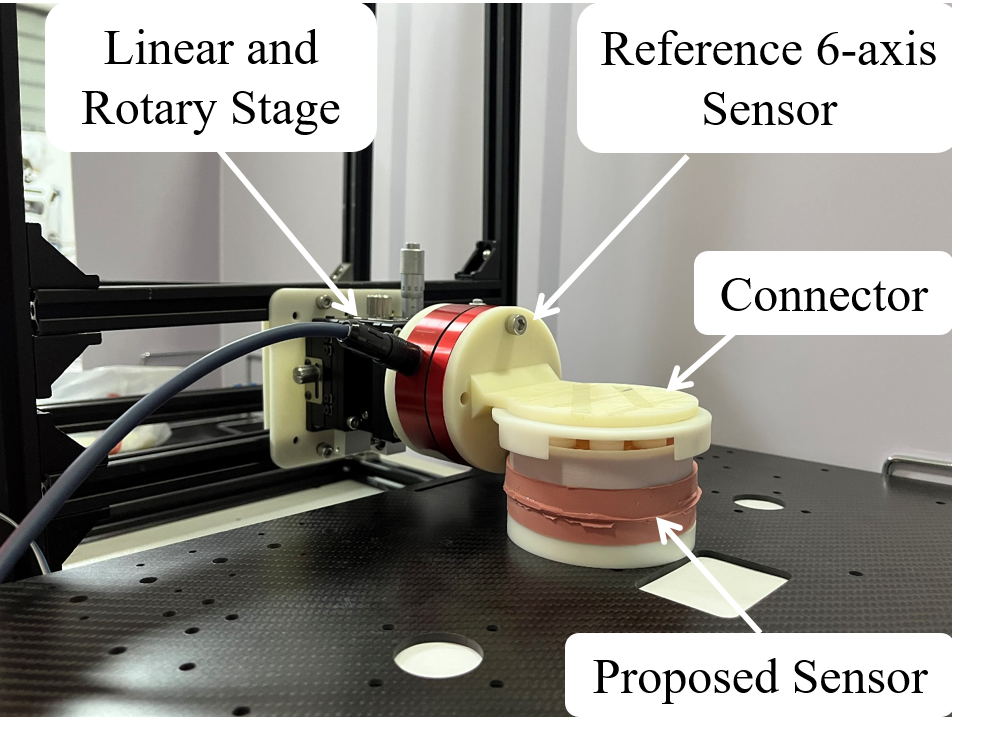}%
\label{b}}
\caption{Calibration and evaluation setup for fabricated prototype sensor.
(a) Setup to apply $F_z$ and $M_z$ to the prototype. (b) Setup to apply $F_x/F_y$ and $T_x/T_y$ to the prototype.}
\label{Setup}
\end{figure}

\subsection{Conversion Law between Force/Torque and Pressure}
If the air in the air-chamber is assumed to be ideal, it follows the ideal gas equation \cite{PGJung}, i.e., the relation between the air pressure and the volume change is linear characteristic, which is,
\begin{equation}
\label{Mt3}
\Delta P_k=-\frac{P_{k0}}{V_{k0}}\Delta V_k=\kappa \Delta V_k,
\end{equation}
where, $P_{k0}, V_{k0}$ are the initial pressure and initial volume of the $k$-$th$ air chamber, $\kappa$ is a constant.

Then we can put forward the conversion matrix based on the above derivation, which are
\begin{align}
\label{Mt4}
\begin{bmatrix} 
F_z\\T_x\\T_y
\end{bmatrix}
&= \mathbf B_l
\mathbf p_l,\\
\begin{bmatrix} 
F_x\\F_y\\T_z
\end{bmatrix}
&= \mathbf B_{u1}
\mathbf p_l
+\mathbf B_{u2}
\mathbf p_u,
\end{align}
where, $\mathbf B_l$, $\mathbf B_{u1}$ and $\mathbf B_{u2}$ are the coefficient matrix as following, which are the inverse functional relation in Eq.~\ref{FT1} and Eq.~2,
\begin{align}
\label{Mt6}
\mathbf B_l&= (\kappa \mathbf T_l)^{-1},\\
\mathbf B_{u1}&=-(\kappa\mathbf T_{u1})^{-1}\mathbf T_{u2} \mathbf T_l^{-1},\\
\mathbf B_{u2}&=(\kappa\mathbf T_{u1})^{-1},
\end{align}
so, the transition law can be expressed as
\begin{align}
\label{Mt7}
\mathbf{F}&=\mathbf{K}\mathbf P,\\
\mathbf{K}&=
\begin{bmatrix}
\mathbf B_{u1} & \mathbf B_{u2}\\
\mathbf B_l & O
\end{bmatrix},
\end{align}
where, $\mathbf{F}=[F_x, F_y, T_z, F_z, T_x, T_y]^T$, $\mathbf{K}\in \mathbb{R}^{6\times16}$ is the transform matrix to be calibrated, $\mathbf{P}=[\mathbf p_l; \mathbf p_u]^T$.

\subsection{Finite Element Model Analysis}
Finite element analysis (FEA) was performed to validate if the theoretical analysis was in line with reality. So the model was imported into FEA simulation software (Abaqus R2018), and the simulation results are as shown in Fig.~S2 in supplementary materials.
The strain energy potential of the elastosis vulcanizing silicone rubber was set to Yeoh, where set $C_{10}=0.11$, $C_{20}=0.02$ \cite{Xavier,Polygerinos}. The density was set to 1130 kg/m$^3$.
The Young's modulus of the aluminum alloy was set to 72 GPa, the Poisson's ratio was set to 0.33. The density was set to 2750 kg/m$^3$\cite{Neila}.

From the Fig.~S2, it can be concluded that the simulation shows the same tendency as the theoretical analysis in section III. 
Take case $b$ as an example, it shows that the volume of the chamber $u_1$ increases while the volume of chamber $u_5$ decreases when the force $F_x$ is loaded.
The other cases followed the same tendency as the theoretical expected above.
\begin{table}[!t]
\caption{Specifications of the Prototype Sensor}
\label{basicinfo}
\centering
\renewcommand\arraystretch{1.2}
\begin{tabular}{c c }
\hline
\hline 
Index & Parameters  \\
\hline
Material &  silicone rubber and Aluminum \\
Size    &  $\Phi$80$\times$20 mm  \\
Weight &  About 100 g \\
Capacity &  50 N (Force), 1 Nm (Torque)\\
Operating Frequency & $<10$ Hz    \\
Average Deviation &   4.9$\%$   \\
Average Repeatability &  2.7$\%$    \\
Average Non-Linearity &  5.8$\%$   \\
Average Hysteresis  &    6.7$\%$   \\
Average Accuracy    &   9.3$\%$  \\
Average Drift &  $\pm$0.83$\%$ \\
Dynamic Response Delay &   3.4 ms\\
Communication Mode & I$^2$C  \\
Recommended Excitation & 1.5-3.6 V  \\
Operating Temp Range &  -40-85 $^\circ$C\\
Cables Specifications & $\Phi$1 mm\\
\hline
\hline 
\end{tabular}
\end{table}

\begin{figure}[!t]
\centering
\subfloat[]{\includegraphics[width=1.7in]{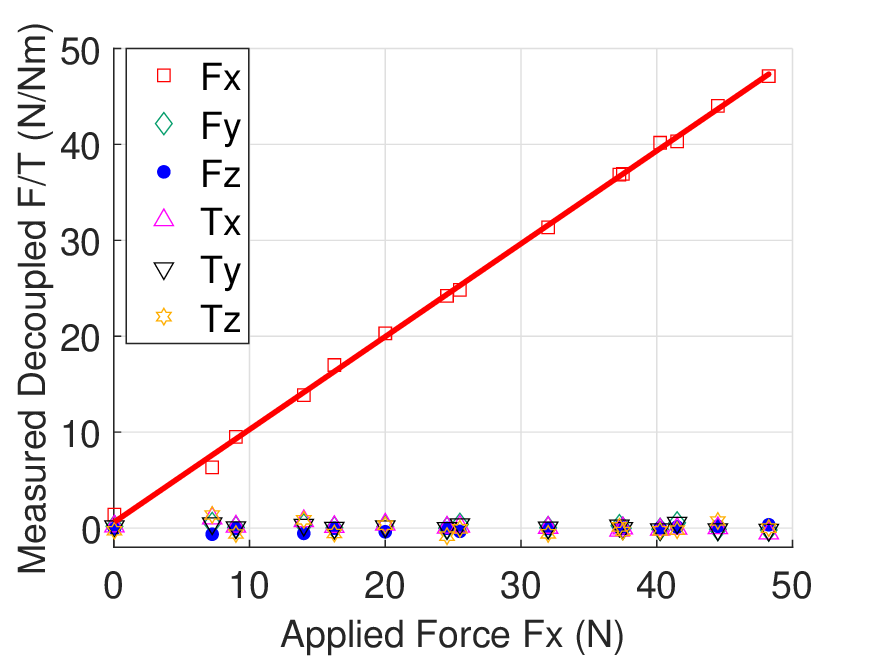}
\label{Fx_st}}
\subfloat[]{\includegraphics[width=1.7in]{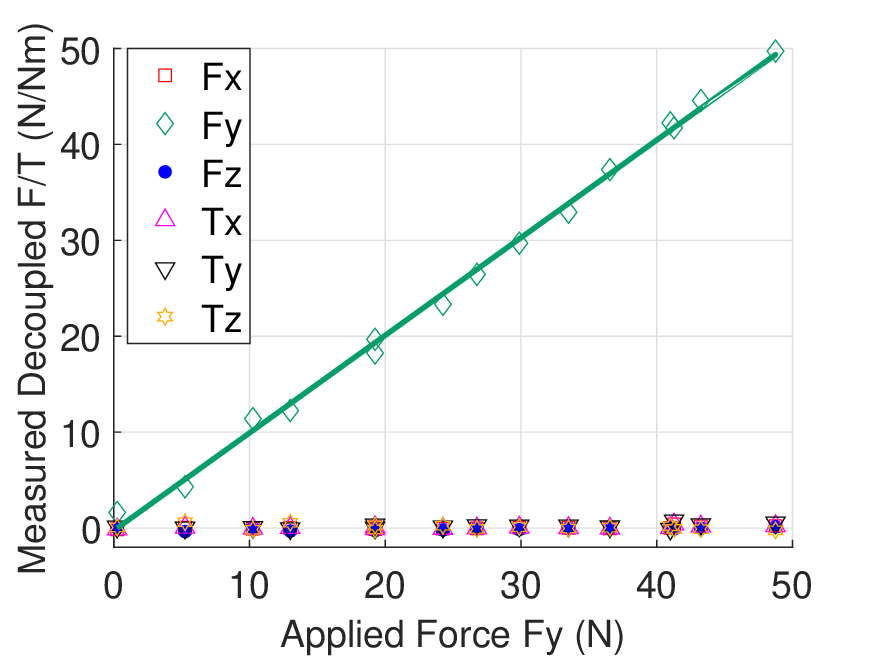}
\label{Fy_st}}
\hfill
\subfloat[]{\includegraphics[width=1.7in]{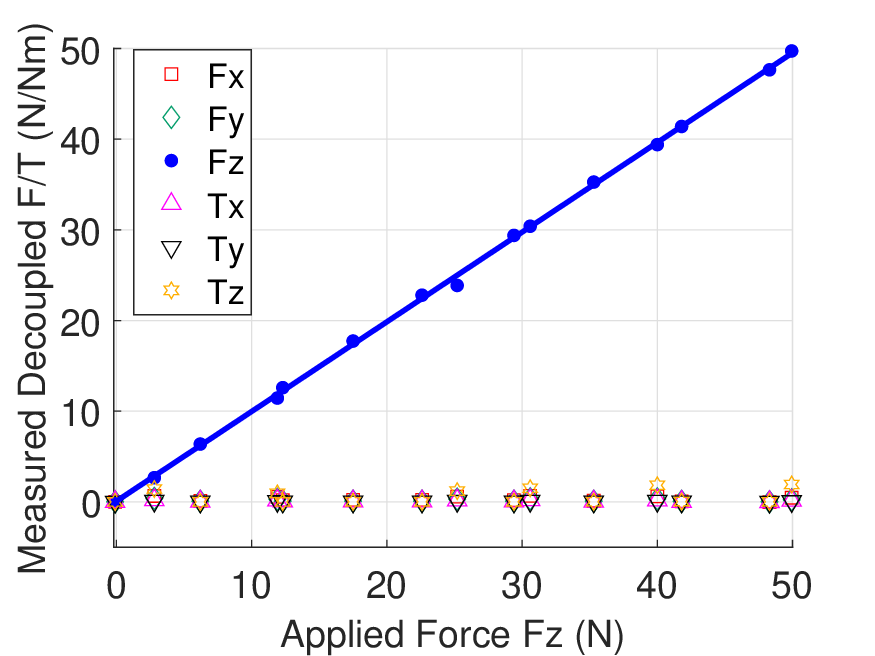}%
\label{Fz_st}}
\subfloat[]{\includegraphics[width=1.7in]{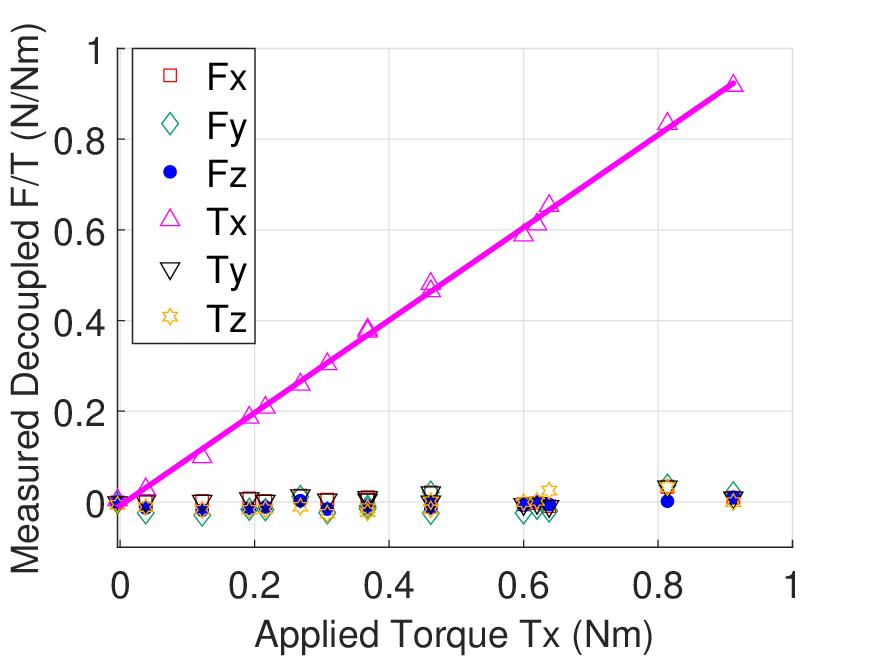}%
\label{Tx_st}}
\hfill
\subfloat[]{\includegraphics[width=1.7in]{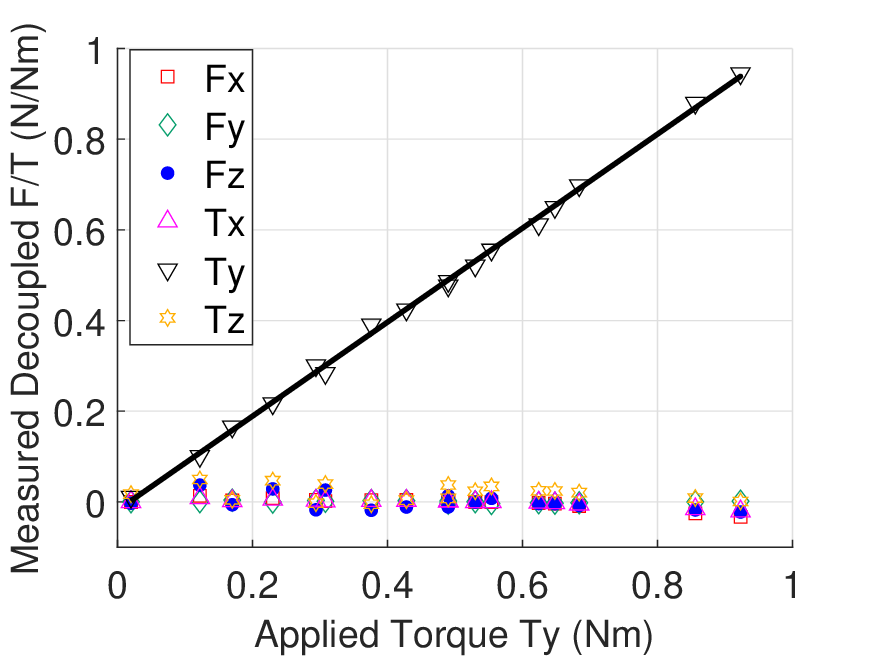}%
\label{Ty_st}}
\subfloat[]{\includegraphics[width=1.7in]{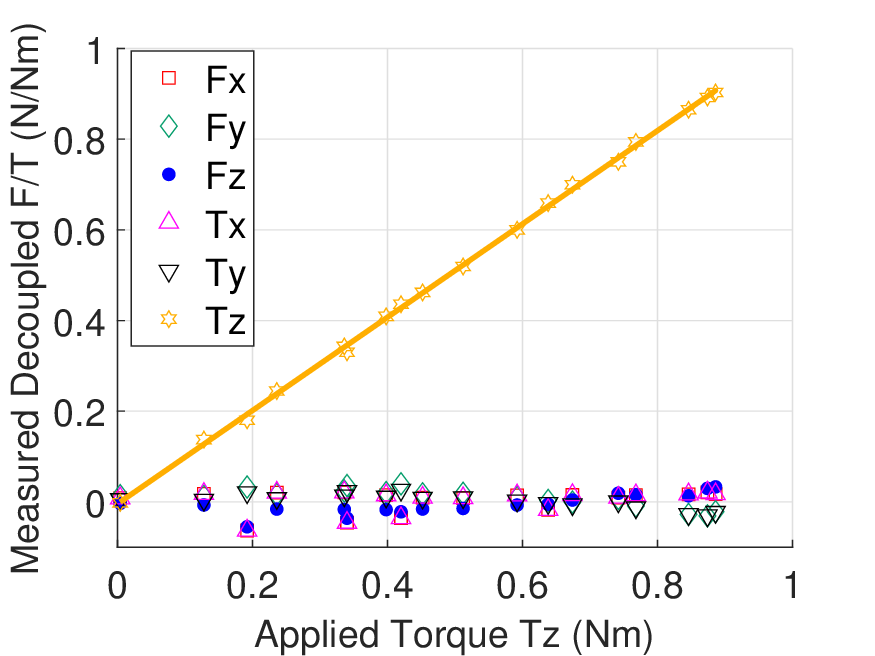}%
\label{Tz_st}}
\caption{Static response of the prototype sensor with respect to the reference sensor in the upload-download way. The slope of fitting line is showed in Tab.~\ref{table1}. (a) $F_x$ (b) $F_y$ (c) $F_z$ (d) $T_x$ (e) $T_y$ (f) $T_z$.}
\label{static2}
\end{figure}

\section{Experiments and Evaluation}
\subsection{Experiment Setup and Calibration}
The prototype sensor and a reference 6-F/T sensor (DynPick-WEF6A, WACOH) are connected in serial using 3D printed material, so the applied force on the prototype and the reference sensor are equivalent, as shown in Fig.~\ref{Setup}. 
The reference 6-F/T sensor features a sensing range of 0-200 N and 0-4 Nm, with corresponding resolutions of 32 LSB/N and 1640 LSB/Nm. It exhibits linearity and hysteresis within 3$\%$ FS and 3$\%$ FS, while other axis sensitivity and drift are $\pm 5\%$ FS and $\pm$0.2$\%$ ${\rm FS/^\circ C}$, respectively.
Force and torque can be applied to the sensors using a rotary stage and a linear stage. The rotary stage and the linear stage are alternatives under different conditions.

In Fig.~\ref{Setup}(a), two loading cases for force or torque are depicted. Case 1, the prototype can sense force $F_z$ when applying a linear movement along the z-axis. Case 2, the prototype can sense torque $T_z$ when applying a rotational movement along the z-axis.
In Fig.~\ref{Setup}(b), two additional loading cases for force or torque are illustrated. Case 3, the prototype can sense force $F_x/F_y$ when applying a linear movement along the z-axis. 
Case 4, the prototype can sense force $T_x/T_y$ when applying a rotational movement along the z-axis.

The goal of the calibration is to determine a transformation matrix $\mathbf{K}$ satisfies Eq.~\ref{Mt7}. The matrix $\mathbf{K}$ is obtained by least square method and the result is in the supplementary materials.

The parameters of the proposed sensor prototype model are shown in Tab.~\ref{basicinfo}.

\begin{table}[!t]
\caption{Static Load Response Regression Equation of the Prototype}
\label{table1}
\centering
\renewcommand\arraystretch{1.2}
\begin{tabular}{c c c c}
\hline
\hline
Applied F/T & \makecell{Regression Equation \\ (Zero-intercept)} & \makecell{Confidence Interval \\ (mean$\pm$std)}  & R$^2$ \\
\hline 
$F_x$ & $y=0.9857x$ & 0.9857$\pm$0.011 & 0.9983\\
$F_y$ & $y=1.010x$ & 1.010$\pm$0.016 & 0.9966\\
$F_z$ & $y=0.9911x$ & 0.9911$\pm$0.008 & 0.9994\\
$T_x$ & $y=1.008x$ & 1.008$\pm$0.013 & 0.9981\\
$T_y$ & $y=1.006x$ & 1.006$\pm$0.014 & 0.9971 \\
$T_z$ & $y=1.022x$ & 1.022$\pm$0.007 & 0.9992\\
\hline
\hline
\end{tabular}
\end{table}

\subsection{Static Load Response}
By using the experiment platform shown in Fig.~\ref{Setup}, the static load response of the prototype was obtained. Fig.~\ref{static2} shows the response of each axis when a single axis force or torque is gradually applied in the upload-download way.
The static load response regression equation can be obtained by linear regression in zero intercept way. The ideal regression coefficient in Tab.~\ref{table1} is 1. The static response regression equation, regression coefficient confidence interval and regression R-square value of the prototype is shown in Tab.~\ref{table1}.

In Fig.~\ref{static2}, it is evident that each axis maintains a highly linear correlation in the static load response, while other axes exhibit minimal disturbance. This demonstrates the superiority of decoupling.

\begin{figure*}[!t]
\centering
\subfloat[]{\includegraphics[width=2.4in]{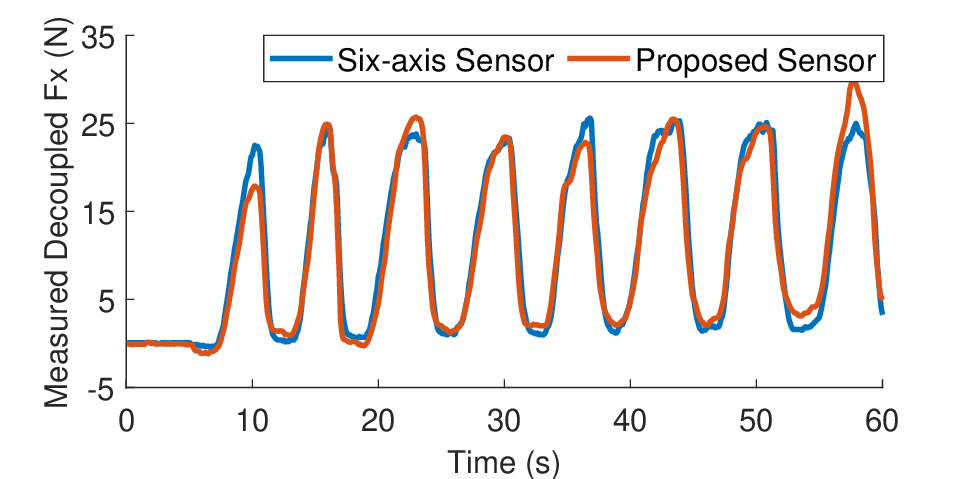}
\label{Fx}}
\subfloat[]{\includegraphics[width=2.4in]{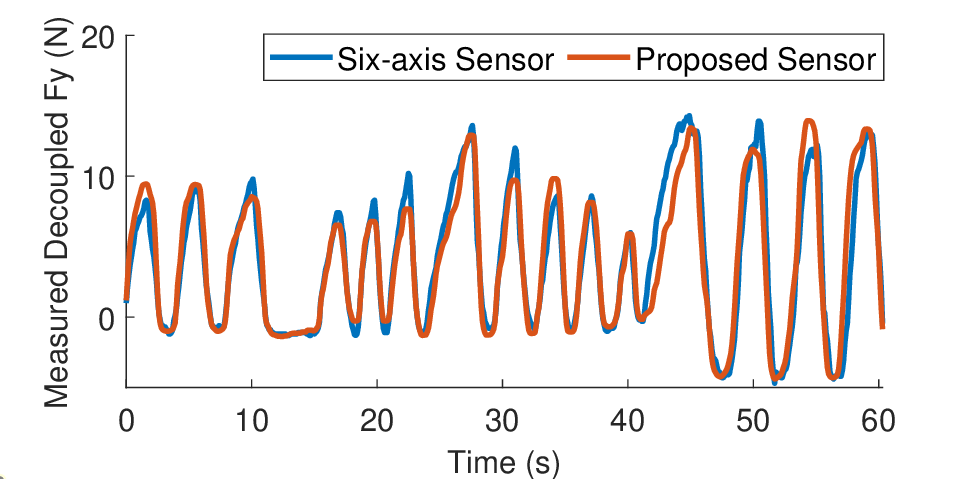}
\label{Fy}}
\subfloat[]{\includegraphics[width=2.4in]{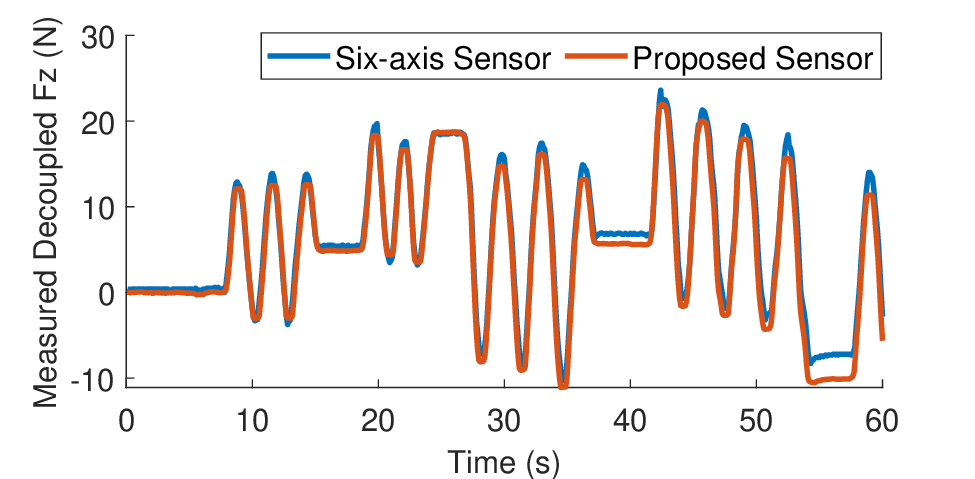}%
\label{Fx}}
\hfil
\subfloat[]{\includegraphics[width=2.4in]{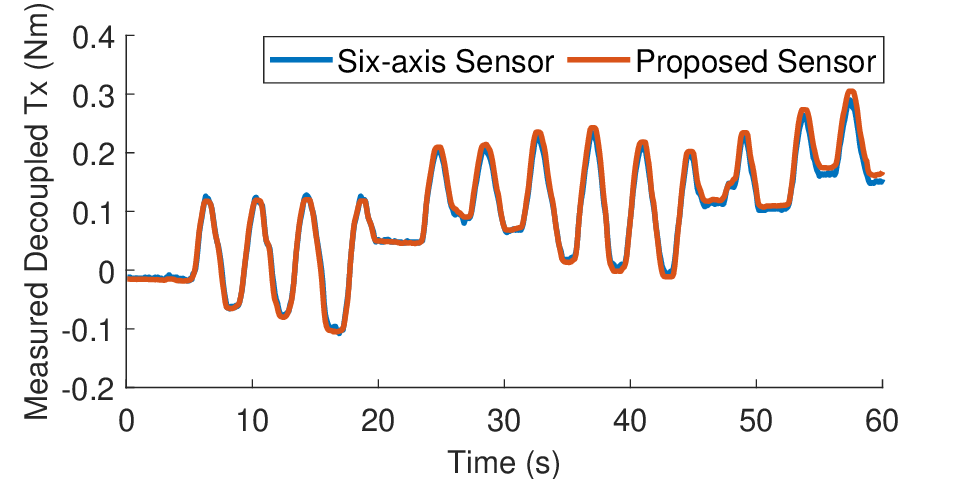}%
\label{Tx}}
\subfloat[]{\includegraphics[width=2.4in]{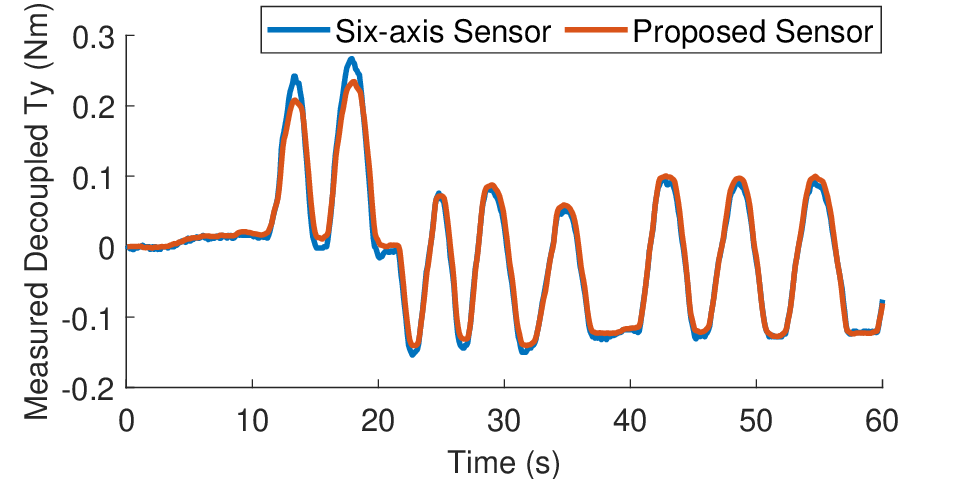}%
\label{Ty}}
\subfloat[]{\includegraphics[width=2.4in]{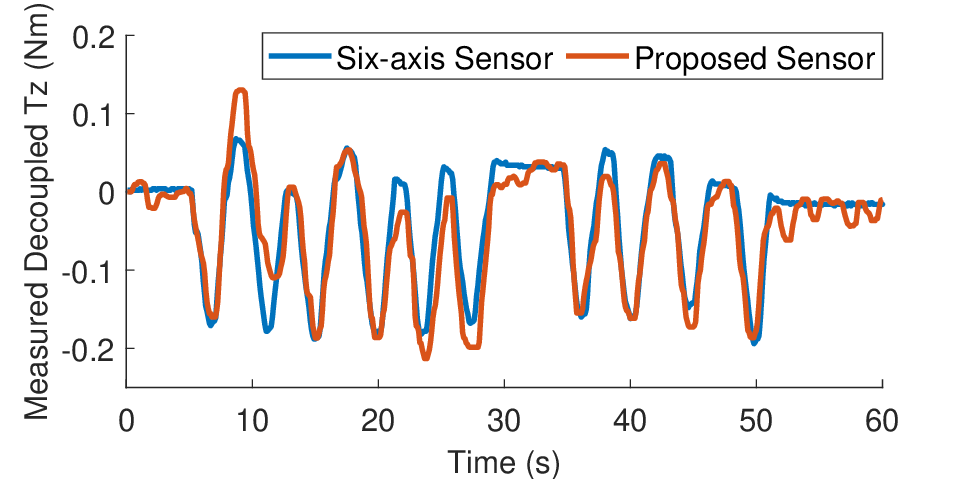}%
\label{Tz}}
\caption{Dynamic load response of the prototype (red) with respect to the reference sensor (blue). (a) $F_x$ (b) $F_y$ (c) $F_z$ (d) $T_x$ (e) $T_y$ (f) $T_z$.}
\label{dy1}
\end{figure*}

\subsection{Dynamic load Response}
Fig.~\ref{dy1} shows the single axis dynamic load time response of the prototype with respect to the reference 6-axis F/T sensor during about one minute. The prototype and the reference sensor are fixed in serial as Fig.~\ref{Setup}. The experiment was performed by applying a random force/torque manually.
The experimental results show that the overall response were pretty matched. 

We repeated the experiment 5 times, each time for 2 minutes.
The dynamic performance of the prototype is shown in Tab.~\ref{table2}.
We choose some quantitative indicators, whose calculation formula are as follows. 
\begin{table}[!t]
\caption{Dynamic Load Response of the Prototype}
\label{table2}
\centering
\renewcommand\arraystretch{1.2}
\resizebox{\columnwidth}{!}{
\begin{tabular}{c c c c c c c}
\hline
\hline
 & $E_{dev}$[$\%$] & $E_{rep}$[$\%$] & $E_{nlin}$[$\%$] & $E_{hys}$[$\%$] & $E_{dri}$[$\%$] & $E_{acc}$[$\%$]\\
\hline
$F_x$ & 5.5$\pm$1.2 & 2.7$\pm$0.5 & 5.7$\pm$0.9 & 7.6$\pm$1.5 & $\pm$1.49& 9.9$\pm$1.8\\
$F_y$ & 6.2$\pm$1.3 & 3.1$\pm$0.5 & 6.2$\pm$1.0 & 8.2$\pm$1.4 & $\pm$1.37 & 10.6$\pm$1.7\\
$F_z$ & 3.8$\pm$0.8 & 2.1$\pm$0.4 & 4.9$\pm$0.8 & 3.5$\pm$1.2 & $\pm$1.14 & 6.3$\pm$1.4\\
$T_x$ & 2.3$\pm$0.6 & 1.9$\pm$0.4 & 4.4$\pm$1.0 & 3.8$\pm$1.1 & $\pm$0.04 & 6.1$\pm$1.5\\
$T_y$ & 2.2$\pm$0.6 & 1.7$\pm$0.3 & 3.7$\pm$0.9 & 4.2$\pm$1.2 & $\pm$0.05 & 5.8$\pm$1.5\\
$T_z$ & 9.2$\pm$1.4 & 4.6$\pm$0.9 & 10.1$\pm$1.2 & 13.1$\pm$2.3 & $\pm$0.08 & 17.0$\pm$2.6\\
\hline
\hline
\end{tabular}}
\end{table}

\subsubsection{Deviation rate $E_{dev}$}
\begin{equation}
\label{Acc}
E_{dev} = \frac{\sum_{t=1}^N |F_i(t)-F_{ref}(t)|}{N \mathrm{max} |F_{ref}(t)|},
\end{equation}
where, $F_i(t), F_{ref}(t)$ refer to the measurement value of the prototype and the reference measurement value of the reference six-axis sensor respectively. $t$ is the time index, $N$ is the total number of samples.

\subsubsection{Repeatability Error $E_{rep}$}
\begin{equation}
\label{Rep}
E_{rep} = \frac{1}{2N} \sum_{k=1}^N \frac{|F_i(k)-F_{ref}(k)|}{\mathrm{max} |F_{ref}(k)|},
\end{equation}
where, $k$ is the sequence index.

\subsubsection{Non-Linearity $E_{nlin}$}
\begin{equation}
\label{lin}
E_{nlin} = \frac{\sum_{t=1}^N |F_i(t)-(\lambda_iF_{ref}(t)+\mu_i)|}{\mathrm{max} |(\lambda_iF_{ref}(t)+\mu_i)|},
\end{equation}
where, $\lambda$ and $\mu$ are constants obtained by linear regression of the measurement from the i-th axis.

\subsubsection{Maximum Hysteresis $E_{hys}$}
\begin{equation}
\label{Hys}
E_{hys} = \frac{1}{2} \frac{\mathrm{max} |F_{i,l}(t)-F_{i,u}(t)|}{\mathrm{max} |F_{i}(t)|},
\end{equation}
where, $F_{i,l}$ and $F_{i,u}$ mean the measurements of proposed sensor under the same applied force respectively when loading and unloading.

\begin{table*}[!t]
\caption{Comparison with Other Sensors}
\label{table3}
\centering
\renewcommand\arraystretch{1.2}
\begin{tabular}{c c c c c c c c c c c c c c}
\hline
\hline
 & Channels & Sensing axis& Size & \multicolumn{2}{c}{Sensing range} &  \multicolumn{4}{c}{Deviation Rate}  & Softness\\
 &   No. &No.  & [mm]   & Force [N] & Torque [Nm] & $F_x/F_y$[$\%$] & $F_z$[$\%$] & $T_x/T_y$[$\%$] &$T_z$[$\%$] & Y/N\\
\hline
Proposed & $\bold {16}$ & $\bold 6$ & 80$\times$20 & $\bold {0{\rm -}50}$ & $\bold {0{\rm -}1}$ & 5.5 & 3.8 & $\bold {2.3} $ & 9.2 & $\bold Y$\\
Choi et al. \cite{Choi} & 3 &3 & 40$\times$10 & 0-13 & - & 3.44 & 1.62 & - & - & Y  \\
Dong-Hyuk et al. \cite{DHyuk} & 6 & 6 & 60$\times$20 & 0-20 & 0-1 & 5.5& 2.3 & 4.5 &4.2 & N\\
Yu et al. \cite{ref42} & 4 &3 & 8$\times$2.6 & 0-1.5 & - & 10.68 & 10.68 &- &- & N\\
Chathuranga et al. \cite{ref48} & 3 & 3 & 8$\times$9 & 0-20 &- & $<$10 &$<$5 &- &-  & Y\\
\hline
\hline
\end{tabular}
\end{table*}

\begin{figure}[!t]
\centering
\subfloat[]{\includegraphics[width=1.7in]{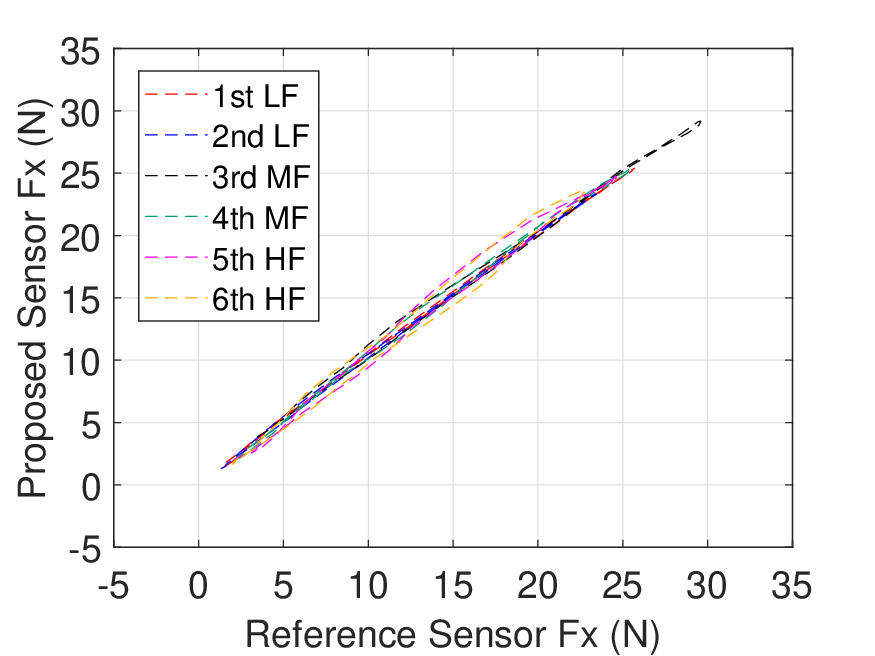}
\label{Fx_rep}}
\subfloat[]{\includegraphics[width=1.7in]{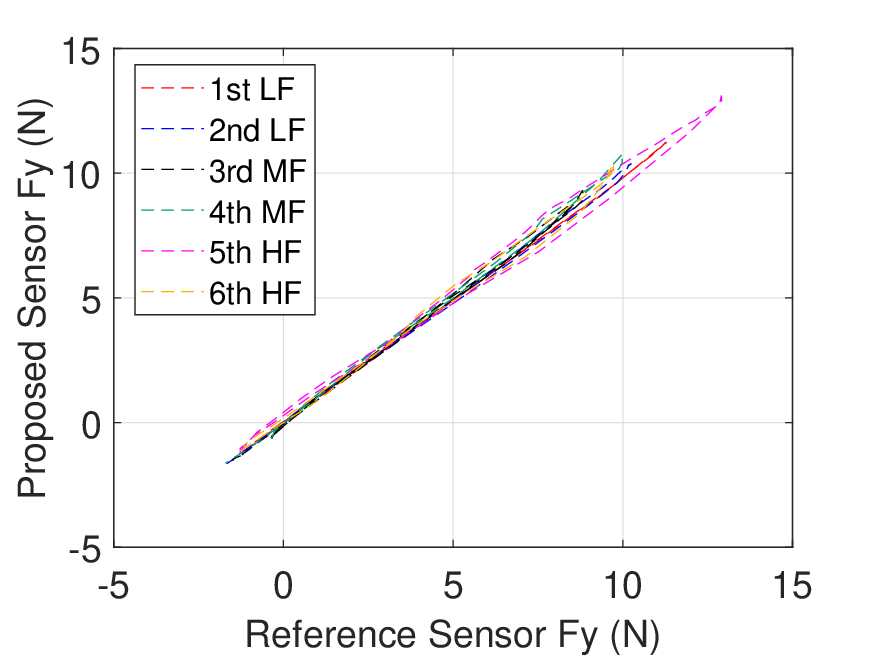}
\label{Fy_rep}}
\hfill
\subfloat[]{\includegraphics[width=1.7in]{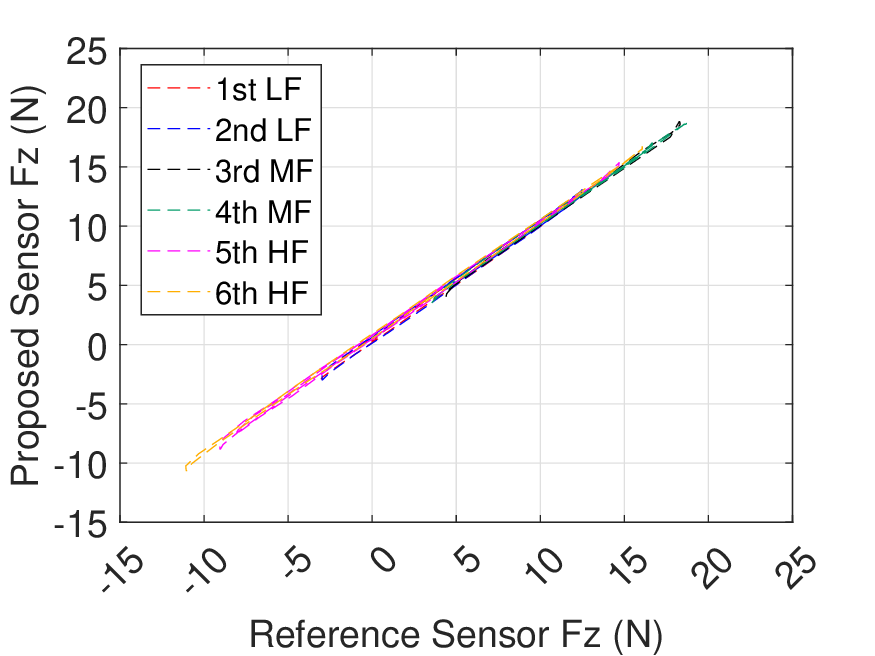}%
\label{Fz_rep}}
\subfloat[]{\includegraphics[width=1.7in]{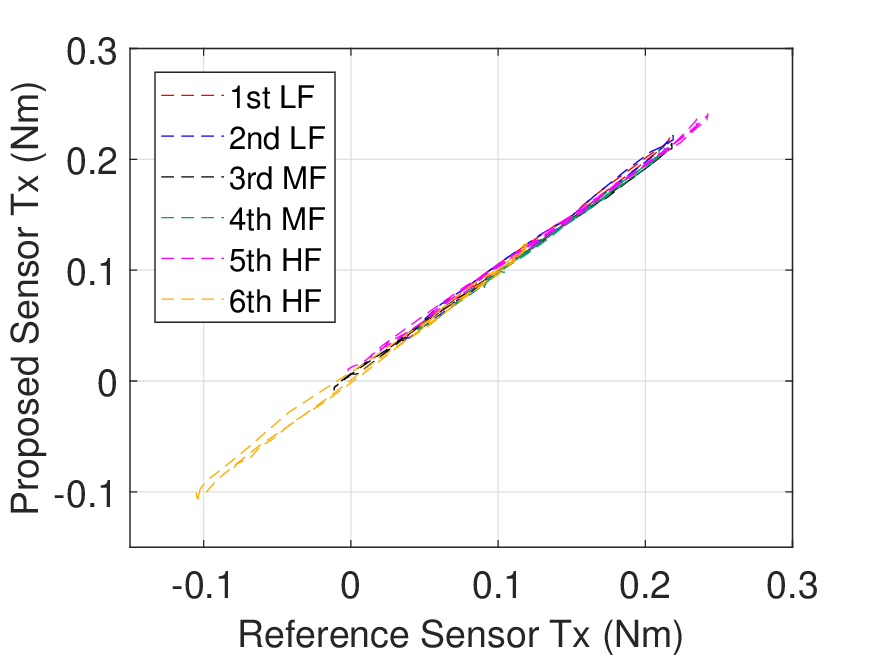}%
\label{Tx_rep}}
\hfill
\subfloat[]{\includegraphics[width=1.7in]{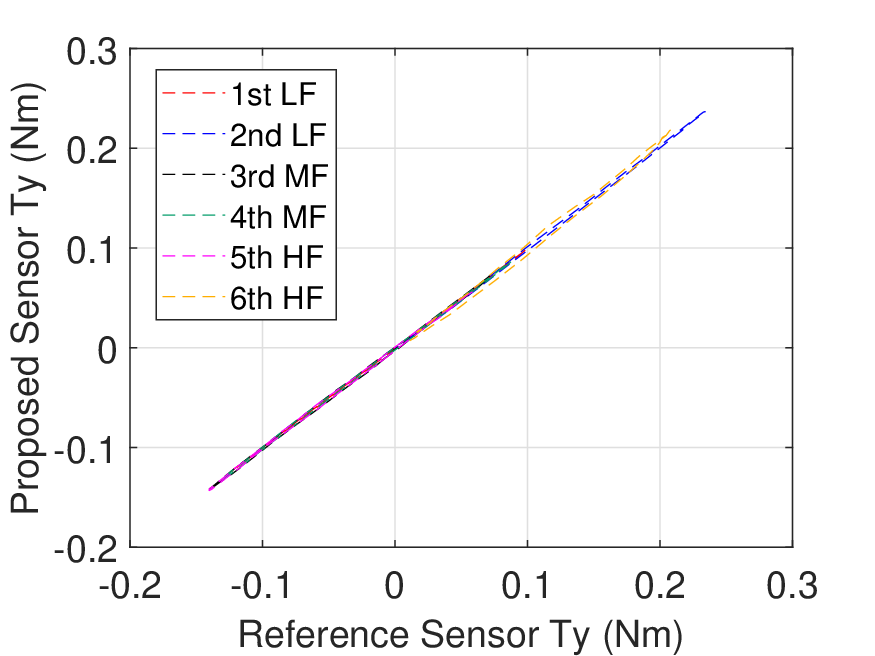}%
\label{Ty_rep}}
\subfloat[]{\includegraphics[width=1.7in]{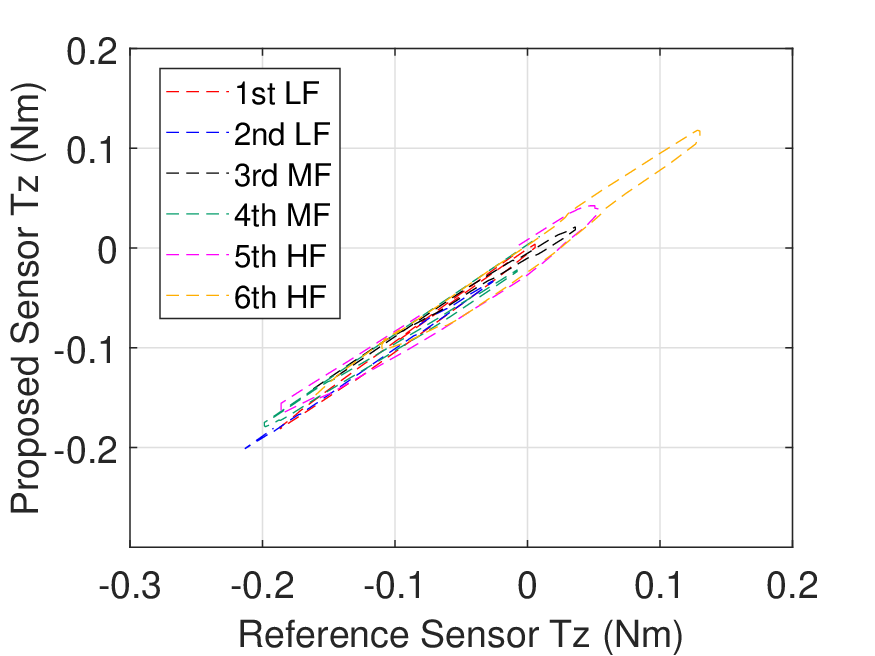}%
\label{Tz_rep}}
\caption{Repeatability and hysteresis response of the prototype in the upload-download way. Three loading speeds under the upload-download way: low frequency (LF, approximately 0.2 Hz), medium frequency (MF, approximately 1 Hz), and high frequency (HF, approximately 5 Hz). (a) $F_x$ (b) $F_y$ (c) $F_z$ (d) $T_x$ (e) $T_y$ (f) $T_z$.}
\label{dy2}
\end{figure}

\subsubsection{Drift Over Time $E_{dri}$}
\begin{equation}
\label{Dri}
E_{dri} = \pm \frac{\sum_{t=1}^N |F_i(t)-F_{ref}|}{N},
\end{equation}
where, $F_{ref}$ refers to the steady state measured value of the reference six-axis sensor without any load, which should be equal to zero.

\begin{figure}[!t]
\centering
\subfloat[]{\includegraphics[width=1.7in]{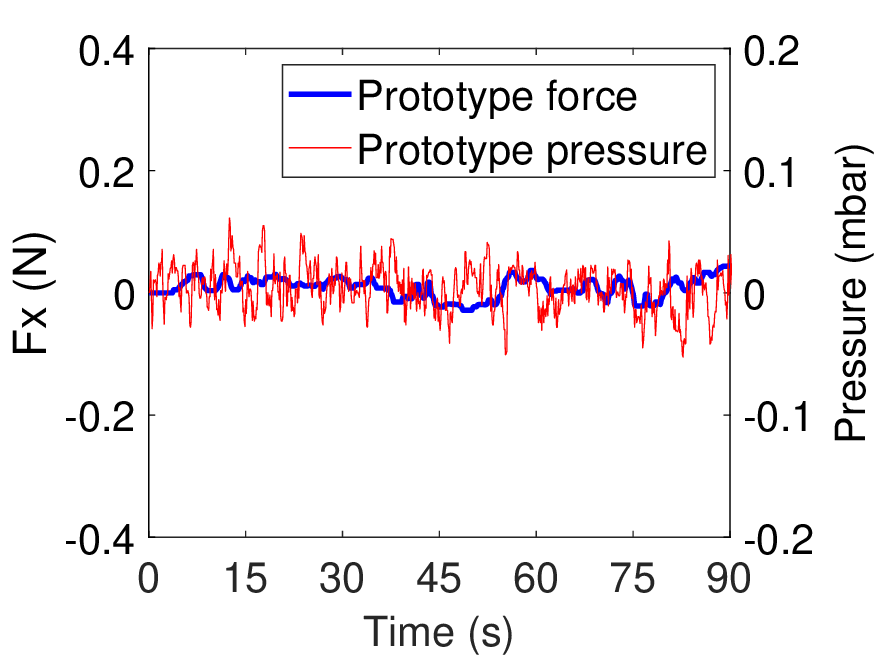}
\label{Fx_dri}}
\subfloat[]{\includegraphics[width=1.7in]{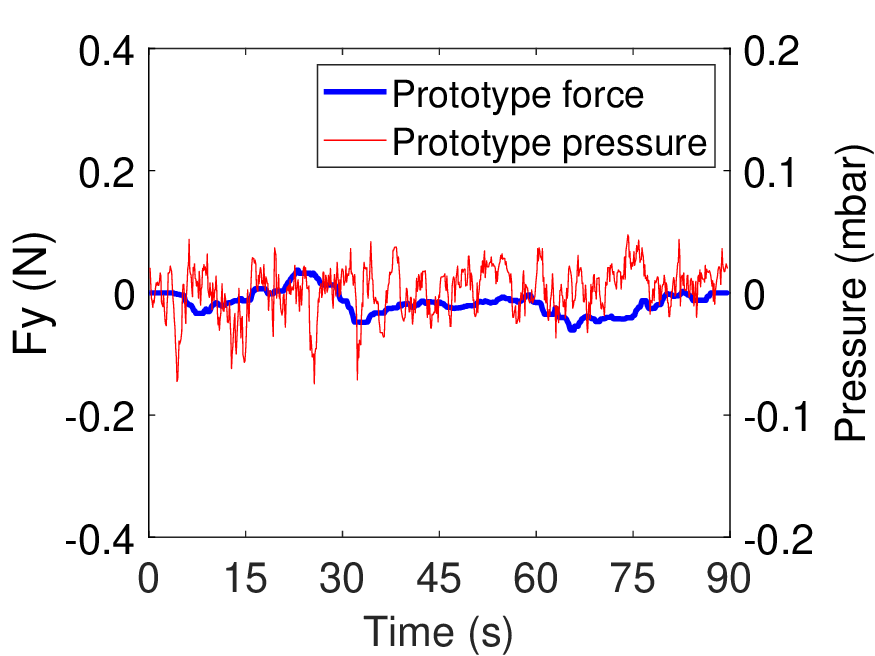}
\label{Fy_dri}}
\hfill
\subfloat[]{\includegraphics[width=1.7in]{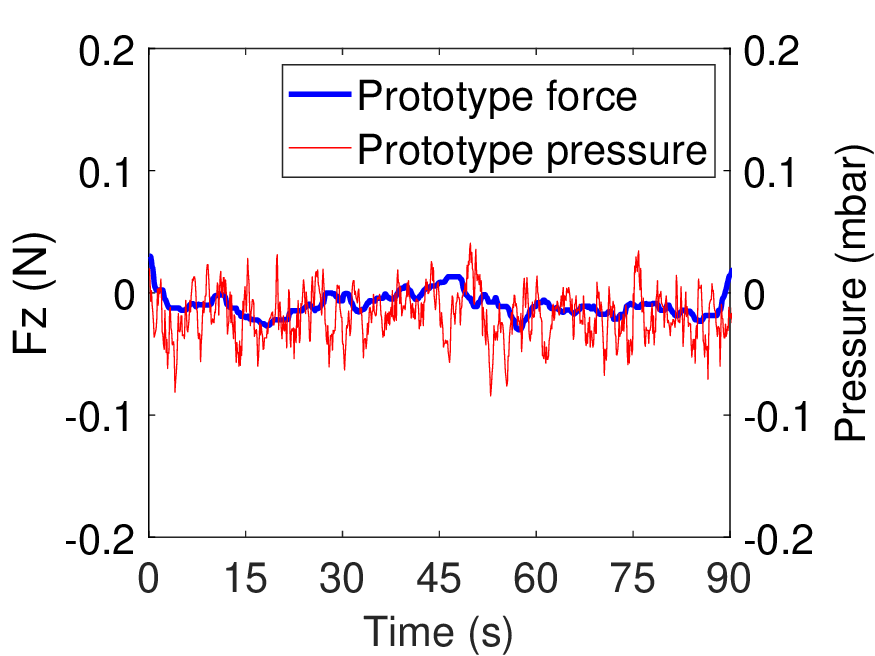}%
\label{Fz_dri}}
\subfloat[]{\includegraphics[width=1.7in]{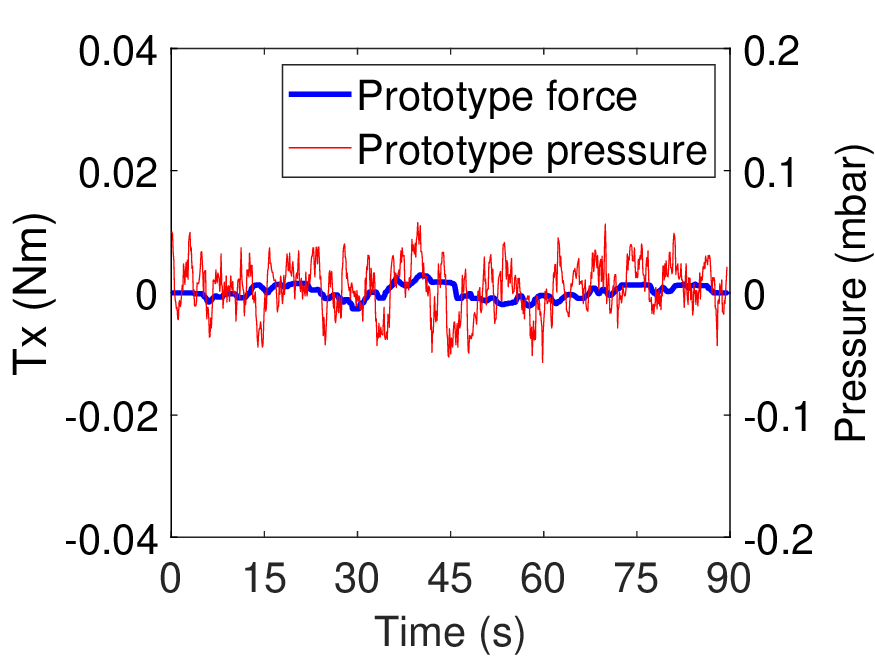}%
\label{Tx_dri}}
\hfill
\subfloat[]{\includegraphics[width=1.7in]{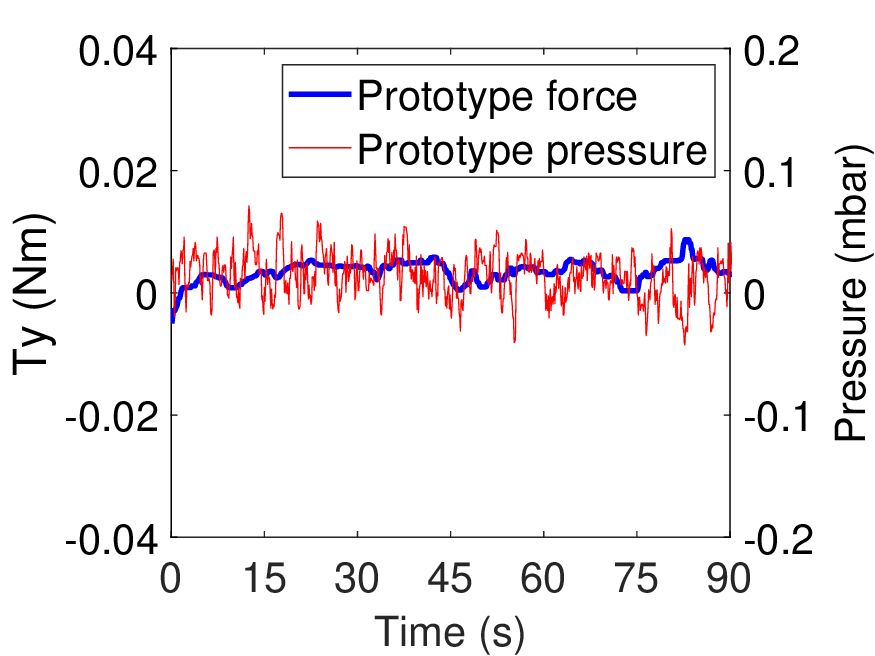}%
\label{Ty_dri}}
\subfloat[]{\includegraphics[width=1.7in]{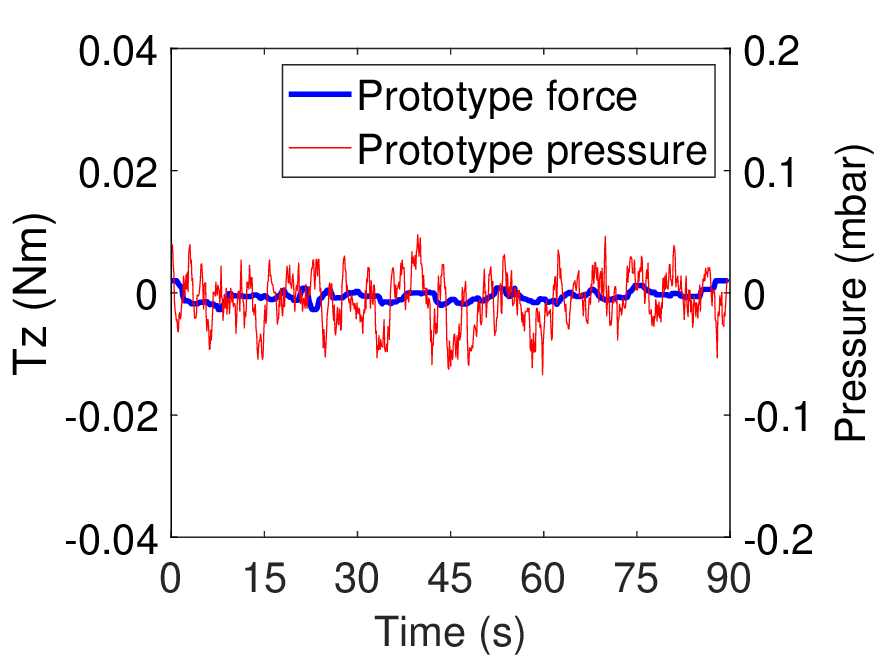}%
\label{Tz_dri}}
\caption{Drift of the prototype (blue) and the drift of the pressure (red). (a) $F_x$ (b) $F_y$ (c) $F_z$ (d) $T_x$ (e) $T_y$ (f) $T_z$.}
\label{dy5}
\end{figure}
\subsubsection{Accuracy $E_{acc}$}
\begin{equation}
\label{Acc}
E_{acc} =  \sqrt{E_{rep}^2+E_{nlin}^2+E_{hys}^2}.
\end{equation}

From the dynamic load response illustrated in Fig.~\ref{dy1}, we can see that the $F_z$, $T_x$ and $T_y$ are eventually converge to the reference values, while the remaining measurements deviate from the reference a little. 
The repeatability and hysteresis characteristics of the prototype were investigated by using the experimental setup shown in Fig.~\ref{Setup}. The experiments were conducted by repeating loading and unloading cycle five times in each direction. The external forces were applied and released gradually while recording the displacement and measured force.
Fig.~\ref{dy2} shows the repeatability and hysteresis response of each axis of the sensor under 6 cycles of loading. Specific data is shown in Tab.~\ref{table2}.
It is evident that the proposed sensor exhibits consistent performance and accuracy across repeatability, non-linearity, and hysteresis indices. Notably, the $F_z$, $T_x$ and $T_y$ axes exhibit excellent performance, while the remaining three axes demonstrate slightly inadequate performance. 
In contrast, traditional force sensors are prone to signal drift over time \cite{DHyuk}. However, given that the measured quantity is air pressure, it is subject to little electrical interference.
Fig.~\ref{dy5} portrays the drift over time in the absence of any load. The drift is minimal across each axis, with the maximum value of $1.49\%$. It is worth noting that the temperature drift characteristics are disregarded since the pressure sensor compensates for them. 
By comparison, the maximum value of the reference commercial sensor is $\pm 0.2\%$ ${\rm FS/^\circ C}$.

\subsection{Dynamic Response Characteristic}
\begin{figure}[!t]
\centering
\includegraphics[width=3in]{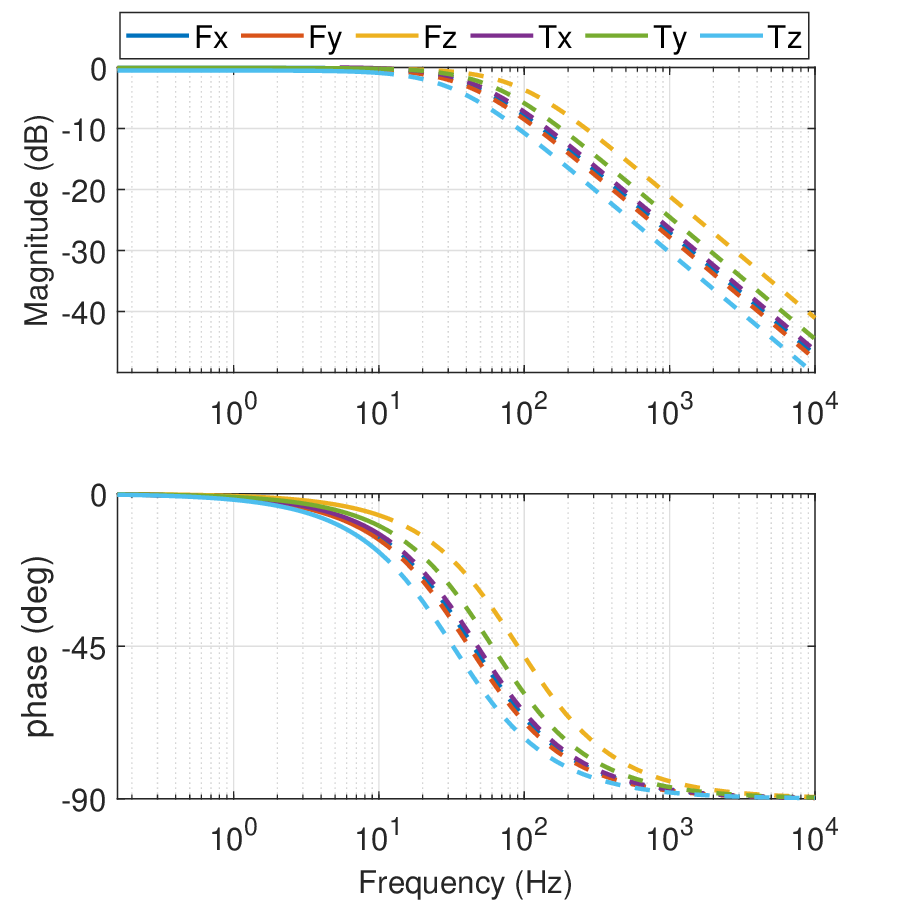}
\caption{Fitted dynamic response curve (Bode diagram) of the prototype. The solid line represents the data bandwidth portion, and the dashed line represents the extrapolation portion.}
\label{Delay}
\end{figure}
Dynamic response characteristic is crucial in the robotics domain. 
Assuming there is little delay in the reference force sensor.
We utilized its measured force signal as the input for the prototype, while the force signal measured by the proposed sensor regarded as the output.
The dynamic response characteristic, represented by the Bode diagram, was obtained for each axis through system identification, as shown in Fig.~\ref{Delay}. The identified transfer functions of the system are presented below,
\begin{equation}
\label{del}
\begin{aligned}
G_{F_x}(s)=\frac{0.9879}{0.0036s+1},
G_{F_y}(s)=\frac{0.9761}{0.0038s+1},\\
G_{F_z}(s)=\frac{0.9815}{0.0018s+1},
G_{T_x}(s)=\frac{1.007}{0.0033s+1},\\
G_{T_y}(s)=\frac{0.9899}{0.0026s+1},
G_{T_z}(s)=\frac{0.9488}{0.0049s+1}.
\end{aligned} 
\end{equation}

The average time constant $\rm \bar T=0.0034$ s. As shown in Fig.~\ref{Delay}, the prototype system maintains good dynamic characteristics in the low frequency range ($<10$ Hz), indicating promising prospects for force measurement applications in the field of robotics. 

\section{Discussions and Limitation}
The sensing performance is influenced differently by rigid and soft materials. In this paper, a rigid-soft hierarchical structure is proposed where the lower layer air chambers are composed of a combination of hard aluminum alloy material and rubber, resulting in good linearity. 
Meanwhile, the other measurements are obtained using silicone rubber. 
The physical models of the pneumatic system are inherently nonlinear and are subject to friction, hysteresis, and threshold behavior, particularly in response to operating pressures \cite{BhabenMMT}. 
The experimental results demonstrate that rigid materials exhibit superior linearity and accuracy compared to soft materials, which exhibit greater non-linearity and reduced accuracy due to factors such as friction and hysteresis.

The main difference between the proposed methodology and the conventional calibration method, under consistent calibration, lies in the decoupling matrix. The dimension of the decoupling matrix (6$\times$16) specified in Eq.~44 implies the necessity to calibrate 96 parameters without direct theoretical analysis. Through the application of a rigid-soft hierarchical structure, the lower-right block of the decoupling matrix is transformed into a zero matrix, resulting in a reduction of the calibrated parameters to 72.
Additionally, detailed in Section III, we established pressure distribution relationships for diverse air chambers under varying loading conditions by analyzing the principles of measurement. In each loading scenario, there is only one unknown parameter. Consequently, only six parameters within the decoupling matrix require calibration.
The advantage of this reduction in calibration parameters is twofold: it simplifies the decoupling relationships and necessitates a smaller dataset for the calibration procedure.

It should be noted that the experiment involves dynamic force loading achieved by turning the knob of the rotating platform, as shown in Fig.~\ref{Setup}. 
The hysteresis performance tests conducted at various loading speeds are depicted in Fig.~\ref{dy2}.
The loading force is not well controlled to be consistent every time, leading to slight variations in the force range during each instance.
This variability results in non-coincident hysteresis loops, which is the main source of non-linearity.
Experiments were carried out in upload-download way at three loading speeds: low frequency (LF, approximately 0.2 Hz), medium frequency (MF, approximately 1 Hz), and high frequency (HF, approximately 5 Hz), enabling an assessment of the hysteresis effect. 
The dynamic loading response diagrams of the test are shown as Fig.~S7 in the supplementary materials.
The hysteresis index $E_{hys}$ under different loading frequency are listed in Tab.~S1 in supplementary materials.
From Tab.~S1, the three axes $F_z$, $T_x$ and $T_y$ show a weak correlation with the loading speed, while the other three axes show a strong correlation. Generally, the results suggest that hysteresis effect becomes more pronounced with faster loading and unloading speeds. However, in practical applications, given that force signals predominantly exhibit low-frequency characteristics and the disparities in loading speeds are not distinctly noticeable, we consider the influence of loading speed on the hysteresis effect to be negligible.

Tab.~\ref{table3} displays a comparison of proposed sensor with others in various research articles. Proposed sensor, compared to others mentioned in the paper, exhibits greater softness and an increased range of six axes. To be specific, it possesses a measuring range of 50 N force and 1 Nm torque. 
The utilization of 6-axis sensing technology necessitates a greater amount of channel information, thereby resulting in an increased size.
The $T_z$ axis shows slightly weaker performance, however, its dynamic load response is substantially effective. This can be attributed to its distinct sensing structure.
It will be a key area of research in the future to enhance measuring accuracy, specifically by augmenting the stiffness difference between two adjacent air chambers.

The proposed sensor has numerous advantages, such as an expanded sensing range and softness. 
However, due to the non-zero permeability of silicone rubber \cite{Choi}, the sensor also has certain limitations. When a continuous large force is applied, air from the air chamber permeates into the material, necessitating a thick outer wall and complex sealing to prevent air leakage, resulting in a larger sensor size. Despite these precautions, complete elimination of permeability is not possible when a large force is continuously applied. 
Hence, mitigating the impact of permeability shall be the forthcoming subject of investigation.

\section{Conclusion}
This paper introduces a soft six-axis force/torque sensor that employs barometers embedded in air chambers using hyper-elastic silicone rubber. 
We design a 16-channel rigid-flexible hierarchical structure to balance the measurement range and accuracy. 
The proposed decoupling method can effectively decouple the six-axis force/torque. 
By employing a rigid-soft hierarchical structure, it simplifies the six-axis decoupling problem into two separate three-axis decoupling problems, thereby reducing the complexity of coupling. As a result, the number of calibration parameters is reduced from 96 to 6.
The effectiveness of the proposed method is demonstrated through finite element model simulations and experiments, which also quantitatively analyze the sensing performance and compare it with other research results. 
The proposed six-axis force/torque (6-F/T) sensor offers a tri-axial force measurement range of up to 50 N and a tri-axial torque measurement range of up to 1 Nm.
The actual measurement ranges could be even larger and could be verified by securely fixing the soft sensor on the load-cell measurement node. With its sensing range and softness, the proposed method has potential applications in various human-machine interaction scenarios, including wearable sensors for measuring human reaction force.

\section*{Acknowledgments}
This work is supported by the National Natural Science Foundation of China (Grant No. U1913207), the International Cooperation Key Program of Hubei Province (No. 2021EHB003) and the Program for HUST Academic Frontier Youth Team.


 
%




\begin{IEEEbiography}[{\includegraphics[width=1in,height=1.25in,clip,keepaspectratio]{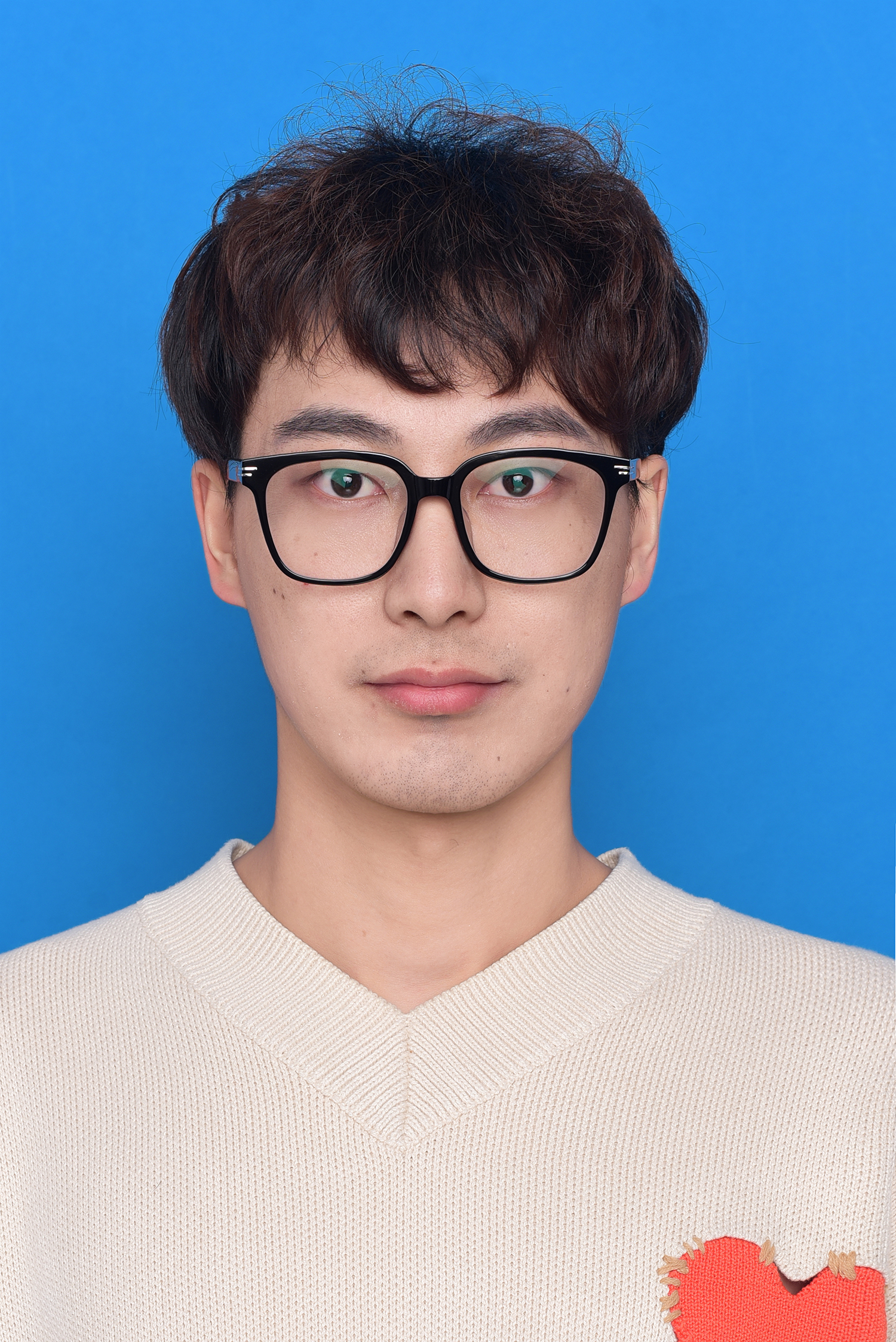}}]{Jun Huo}
received the B.S. degree in mechanical engineering from Northeastern University, Shenyang, China, in 2018. He is currently working toward the Ph.D. degree in control science and engineering of Huazhong University of Science and Technology. His current research interests include mechanical design, sensing and control of rehabilitation robot. 
\end{IEEEbiography}

\begin{IEEEbiography}[{\includegraphics[width=1in,height=1.25in,clip,keepaspectratio]{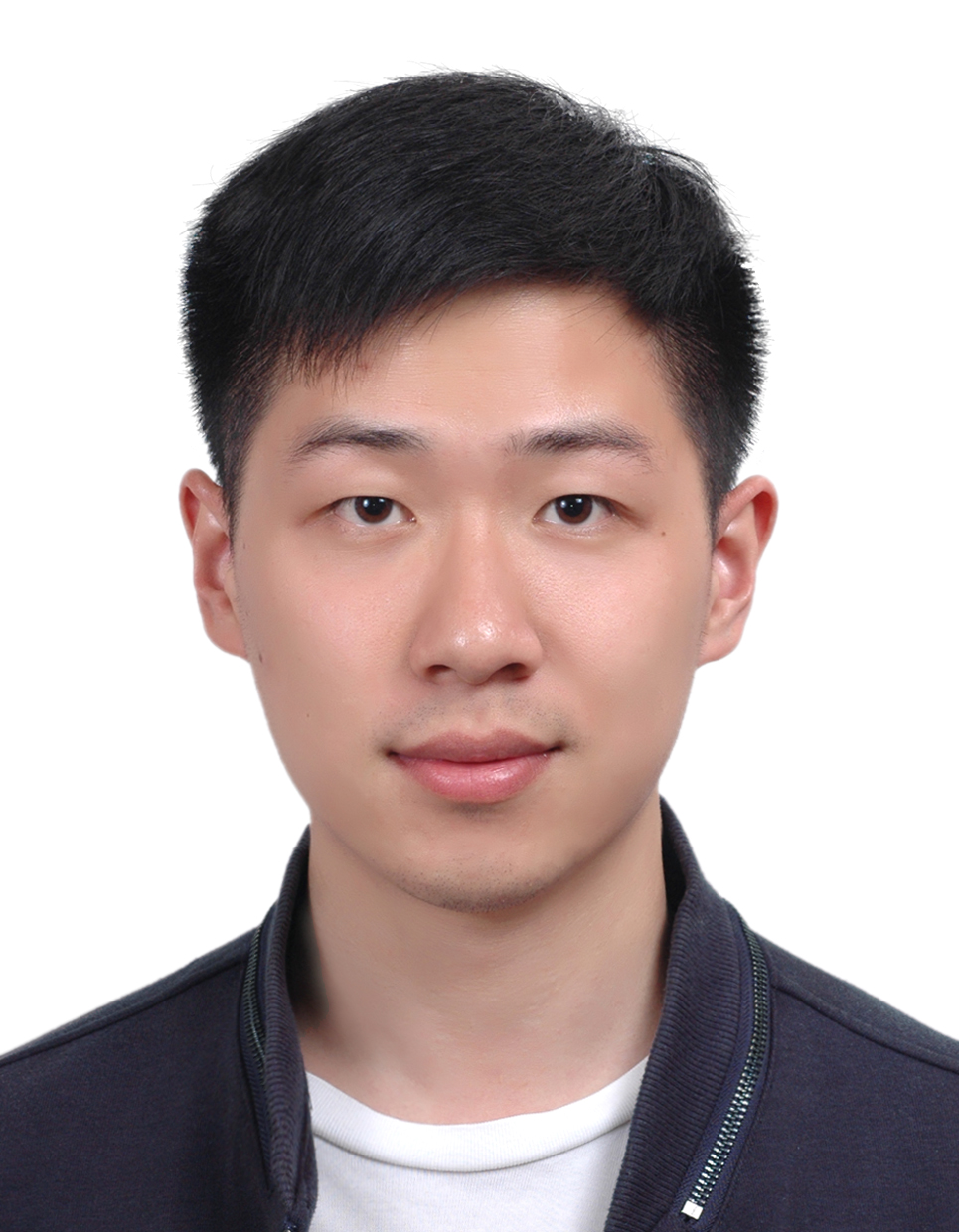}}]{Hongge Ru}
received the B.S. degree in control science and engineering from Tongji University, Shanghai, China, in 2015, and now is a Ph.D. candidate in control science and engineering of Huazhong University of Science and Technology. His current research interests include soft robotics, modelling and identification of bending/linear pneumatic muscle, hysteresis, design and control of supernumerary robotic limbs based on pneumatic muscle.
\end{IEEEbiography}

\begin{IEEEbiography}[{\includegraphics[width=1in,height=1.25in,clip,keepaspectratio]{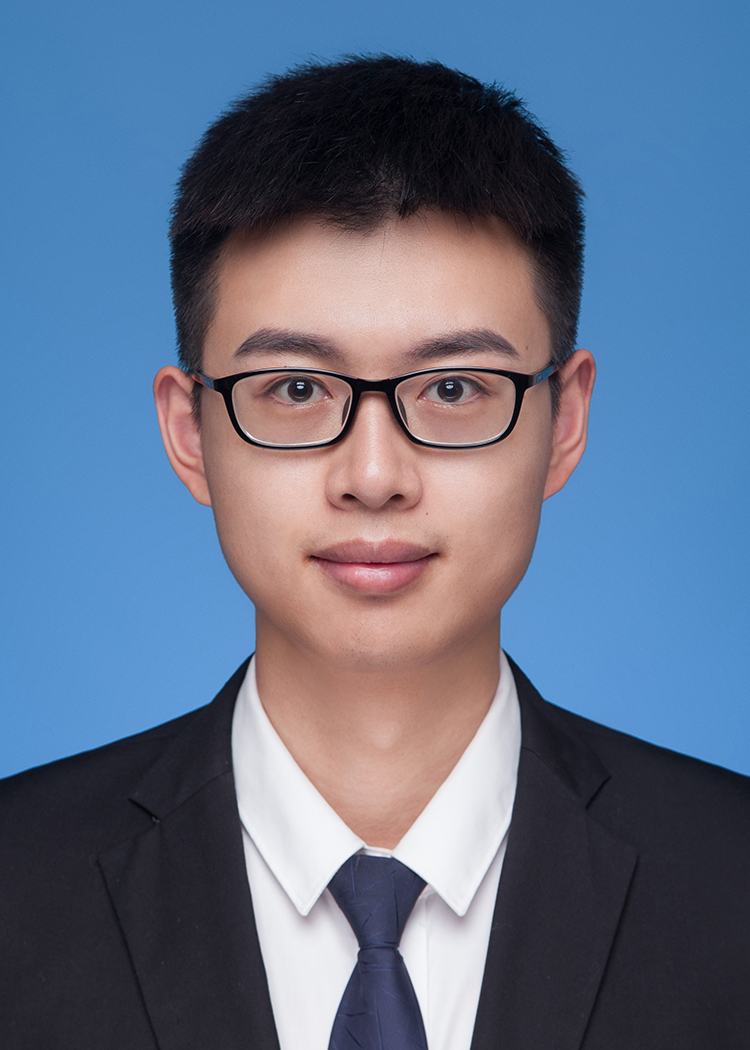}}]{Bo Yang}
received the B.S. degree in automation from Chongqing University, Chongqing, China, in 2019, and now is a Ph.D. candidate in control science and engineering of Huazhong University of Science and Technology. His current research interests include human intention recognition, intention-based robot control, and set-membership filter.
\end{IEEEbiography}

\begin{IEEEbiography}[{\includegraphics[width=1in,height=1.25in,clip,keepaspectratio]{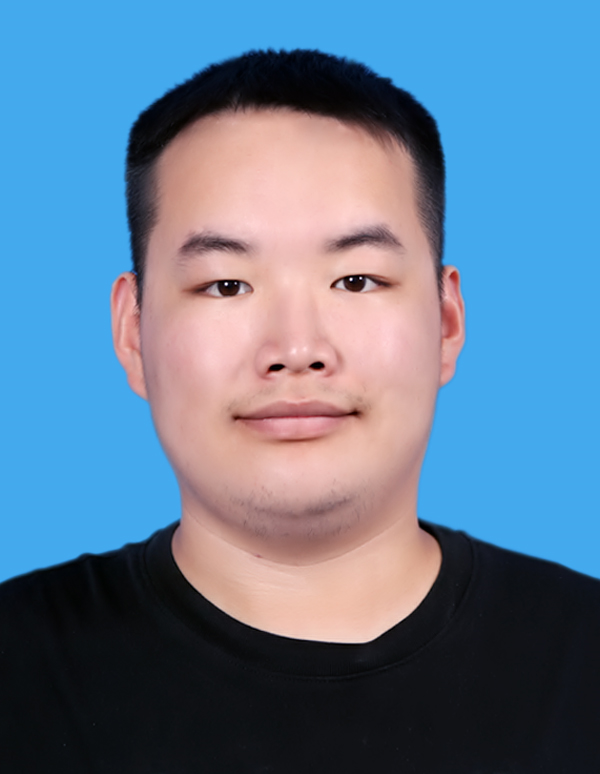}}]{Xingjian Chen}
received the B.S. degree in artificial intelligence automation from Huazhong University of Science and Technology, Wuhan, China, in 2022, and now is a master candidate in artificial intelligence automation of Huazhong University of Science and Technology. His current research interests include soft robotics, multi-task learning, design and control of lower limb robotics based on multi-mode sensors.
\end{IEEEbiography}

\begin{IEEEbiography}[{\includegraphics[width=1in,height=1.25in,clip,keepaspectratio]{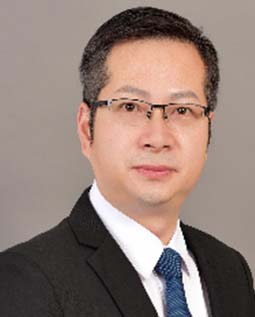}}]{Xi Li}
received the Ph.D. degree from Shanghai Jiao Tong University, Shanghai, China, in 2006. From 2011 to 2012, he was a Visiting Scholar with the Naval Architecture and Marine Engineering Department, University of Michigan, Ann Arbor, MI, USA. He has been with the Huazhong University of Science and Technology, Wuhan, China, since 2006, where he is currently a Professor with the School of Artificial Intelligence and Automation. His research interests include intelligent control and model-predictive control of renewable energy systems.
\end{IEEEbiography}

\begin{IEEEbiography}[{\includegraphics[width=1in,height=1.25in,clip,keepaspectratio]{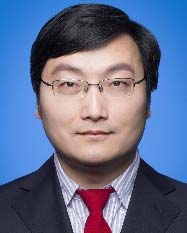}}]{Jian Huang}
received the Graduate, M.E., and Ph.D. degrees in control science and control engineering from the Huazhong University of Science and Technology (HUST), Wuhan, China, in 1997, 2000, and 2005, respectively. He is currently a Full Professor with the School of Artificial Intelligence and Automation, HUST. His main research interests include rehabilitation robots, robotic assembly, and assistive robots.
\end{IEEEbiography}




\vfill

\end{document}